\documentclass[10pt,journal,compsoc]{IEEEtran}
\usepackage[hidelinks,colorlinks=false,breaklinks=true,pagebackref=false,allcolors=black]{hyperref}
\usepackage{url}
\usepackage{ragged2e}
\usepackage{makecell}
\usepackage{amsfonts}
\usepackage{amsmath}
\usepackage{amsthm}
\usepackage{graphicx}
\usepackage{algorithm}
\usepackage{subfig}
\usepackage{multirow}
\usepackage{booktabs}
\usepackage{stfloats}
\usepackage{color}
\usepackage{xcolor}
\usepackage{algorithm}
\usepackage{algorithmicx}
\usepackage{algpseudocode}
\usepackage{amsmath}

\usepackage[nocompress]{cite}
\ifCLASSINFOpdf
\else
\fi
\begin{document}
%
\title{Interaction Pattern Disentangling for Multi-Agent Reinforcement Learning}
%
%
%
%

\author{Shunyu~Liu,
        Jie~Song,
        Yihe~Zhou,
        Na~Yu,
        Kaixuan~Chen,
        Zunlei Feng,
        Mingli~Song
\thanks{
  This article has been accepted for publication by IEEE Transactions on Pattern Analysis and Machine Intelligence. The published version is available at \url{https://doi.org/10.1109/TPAMI.2024.3399936}. This work was supported in part by the Joint Funds of the Zhejiang Provincial Natural Science Foundation of China under Grant LHZSD24F020001, in part by the Zhejiang Province ``LingYan" Research and Development Plan Project under Grant 2024C01114, and in part by the Zhejiang Province High-Level Talents Special Support Program ``Leading Talent of Technological Innovation of Ten-Thousands Talents Program" under Grant 2022R52046.
  \textit{(Corresponding author: Jie Song.)}
  }
\IEEEcompsocitemizethanks
{\IEEEcompsocthanksitem Shunyu Liu, Yihe Zhou, Na Yu, Kaixuan Chen, and Mingli Song are with the State Key Laboratory of Blockchain and Security, Zhejiang University, Hangzhou 310027, China, and also with the Hangzhou High-Tech Zone (Binjiang) Institute of Blockchain and Data Security, Hangzhou 310027, China (e-mail: liushunyu@zju.edu.cn; yihe\_zhou@zju.edu.cn; na\_yu@zju.edu.cn; chenkx@zju.edu.cn; brooksong@zju.edu.cn).
\IEEEcompsocthanksitem Jie Song and Zunlei Feng are with the School of Software Technology, Zhejiang University, Hangzhou 310027, China (e-mail: sjie@zju.edu.cn; \text{zunleifeng@zju.edu.cn}).
}
\thanks{Digital Object Identifier 10.1109/TPAMI.2024.3399936}
}

%
%


\markboth{IEEE TRANSACTIONS ON PATTERN ANALYSIS AND MACHINE INTELLIGENCE}
{Liu \MakeLowercase{\textit{et al.}}: Interaction Pattern Disentangling for Multi-Agent Reinforcement Learning}

%


\IEEEpubid{\begin{minipage}{\textwidth}\ \\[30pt] \centering 0162-8828~\copyright~2024 IEEE. Personal use is permitted, but republication/redistribution requires IEEE permission.\\See https://www.ieee.org/publications/rights/index.html for more information.\end{minipage}}


\IEEEtitleabstractindextext{%
\begin{abstract}
  \justifying
  Deep cooperative multi-agent reinforcement learning has demonstrated its remarkable success over a wide spectrum of complex control tasks. However, recent advances in multi-agent learning mainly focus on value decomposition while leaving entity interactions still intertwined, which easily leads to over-fitting on noisy interactions between entities. In this work, we introduce a novel \emph{interactiOn Pattern disenTangling}~(OPT) method, to disentangle the entity interactions into interaction prototypes, each of which represents an underlying interaction pattern within a subgroup of the entities. OPT facilitates filtering the noisy interactions between irrelevant entities and thus significantly improves generalizability as well as interpretability. Specifically, OPT introduces a sparse disagreement mechanism to encourage sparsity and diversity among discovered interaction prototypes. Then the model selectively restructures these prototypes into a compact interaction pattern by an aggregator with learnable weights. To alleviate the training instability issue caused by partial observability, we propose to maximize the mutual information between the aggregation weights and the history behaviors of each agent. Experiments on single-task, multi-task and zero-shot benchmarks demonstrate that the proposed method yields results superior to the state-of-the-art counterparts. Our code is available at \url{https://github.com/liushunyu/OPT}.
\end{abstract}

}

\maketitle

\IEEEdisplaynontitleabstractindextext

%
\IEEEpeerreviewmaketitle

\IEEEraisesectionheading{\section{Introduction}\label{sec:introduction}}

\IEEEPARstart{R}{einforcement} Learning~(RL) is instigated by the \textit{trial and error}~(TE) procedure in human behaviors and now becomes a well-established paradigm for building intelligent systems interacting with the environment, especially driven by the resurgence of deep learning~\cite{sutton2018reinforcement,DQN15,AlphaGo,li2021structured,DBLP:journals/pami/AkrourTP22,DBLP:journals/pami/LiPFZXZ23}. In the RL context, cooperative Multi-Agent Reinforcement Learning (MARL), where a group of agents works collaboratively for one common goal, is a long-standing research topic and a key tool used to address many real-world problems such as video games~\cite{AlphaStar,FTW}, traffic light systems~\cite{chu2019multi,wu2020multi}, and smart grid control~\cite{yan2020multi,xu2020multi,wang2021multi}. However, cooperative MARL is perceived as a significantly more challenging problem than the single-agent counterpart due to several peculiar characteristics~\cite{zhang2019multi}, for example, 
\textit{credit assignment} that the agents need to deduce their contributions in the presence of only global rewards, 
\textit{partial observability} which necessitates the learning of decentralized policies, 
the \textit{curse of dimensionality} that the joint action~(and state) space grows exponentially with the number of agents, \textit{etc}.

\begin{figure}[!t]
  \centering
  \includegraphics[width=0.48\textwidth]{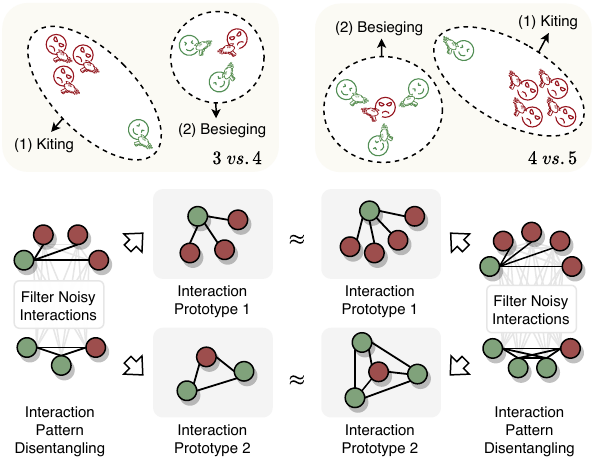}
  \caption{A visualization example of two shooting tasks with different scales. Green and red indicate the agents and enemies, respectively. The intertwined entity interactions can be disentangled into two interaction prototypes by filtering the noisy interactions: (1) One agent kites most of the enemies. (2) The other agents besiege the alone enemy.}
  \label{fig:demo}
\end{figure}

\IEEEpubidadjcol

To alleviate these issues, value-based MARL has recently emerged as a powerful framework
with the large capacity of deep neural networks and the \textit{Centralized Training and Decentralized Execution}~(CTDE) paradigm~\cite{VDN,QMIX,QPLEX,REFIL}. 
Value-based MARL endows agents the ability to learn the individual optimal policies by decomposing the global value function according to their contributions.
For example, Value Decomposition Network~(VDN)~\cite{VDN} learns to additively factorize the team value function into agent-wise value functions, which enables the decentralized deployment of the centralized trained agents. QMIX~\cite{QMIX} identifies that the full factorization of VDN is not necessary and employs a network that estimates joint action values as a complex non-linear combination of per-agent values that condition only on local observations. 
Furthermore, randomized entity-wise factorization for imagined learning~(REFIL)~\cite{REFIL} randomly divided entities into subgroups to construct an auxiliary regularization term, which facilitates discovering more generalized value functions. 
These works mainly focus on agent-wise value decomposition with centralized training for decentralized deployment, while ignoring the specific interconnections between the entities of interest. The decentralized policies, however, can generalize poorly at test time due to over-fitting on noisy interactions during the training time. Specifically, we observe that entity interactions can be disentangled into several interaction prototypes, as depicted in Figure~\ref{fig:demo}. Each interaction prototype corresponds to an underlying interaction pattern within a subgroup of the entities, while different interaction prototypes often simultaneously exist among entities in a complex environment. The interaction prototypes concealed in various-scale tasks are usually shareable, resulting in similar subgroup strategies of agents. Unfortunately, these patterns are closely intertwined and inevitably collapse into a single dense attention distribution in the existing approaches~\cite{MAAC,ATOC,QPLEX,REFIL}. Each agent is trained to pay attention to all observable entities, including the irrelevant ones. These irrelevant entities can be considered as noisy information, severely impeding the extraction of interaction prototypes and downgrading the model generalizability.

In this work, we thus propose \emph{interactiOn Pattern disenTangling}, abbreviated as OPT, to explicitly conduct interaction pattern disentangling for MARL. Unlike prior methods devoted to value decomposition while leaving latent interaction patterns entangled between entities, the proposed OPT disentangles the entity interactions into several \textit{interaction prototypes}. These prototypes reveal the latent behavior semantics to characterize the interaction patterns in different subgroup entities, which facilitates filtering the noisy interactions between irrelevant entities and thus improves the generalizability and interpretability. Technically, we adopt a sparse activation function with a contrastive disagreement objective function to extract both sparse and diverse interaction prototypes. 
To derive the final policy, we selectively recombine these prototypes by an aggregator with learnable weights, unweighting the irrelevant prototypes and emphasizing the important ones. Moreover, we propose to maximize the mutual information between the aggregation weights and the history behaviors to alleviate the training instability issue caused by the partial observability. To sum up, we make the following contributions in this work:
\begin{itemize}
  \item We identify the entangled interaction problem in MARL, and introduce the interaction pattern disentangling task for cooperative MARL, a highly important ingredient for generalization yet largely overlooked by existing literature. 
  \item We propose OPT, a novel disentangling method to decompose entity interactions into interaction prototypes. OPT not only enjoys stronger generalizability, but also exhibits higher interpretability with the disentangled interaction prototypes. 
  \item Experiments conducted on the StarCraft II micromanagement, Google Research Football and Predator-Prey benchmarks demonstrate that OPT yields significantly superior performance to state-of-the-art competitors.
\end{itemize}

The remainder of this manuscript is organized as follows. In Section~\ref{sec:relate}, we review some MARL topics highly related to this work. Then preliminaries of MARL are provided in Section~\ref{sec:background}. The proposed interaction pattern disentangling is provided in Section~\ref{sec:method}. Experimental results, including the benchmark comparison and ablation study, are presented in Section~\ref{sec:result}. Finally, we conclude this manuscript with some possible future works in Section~\ref{sec:conclusion}.

\section{Related Work\label{sec:relate}}

We briefly review here some recent advances that are most related to the proposed work in the context of MARL. 

\subsection{Value-Based MARL}

\textit{Centralized Training and Decentralized Execution}~(CTDE) has recently emerged as a powerful paradigm in MARL, where agents are granted access to additional global information during centralized training and deliver actions only based on their local observation in a decentralized way.
The most common approach to the MARL problem heretofore is to learn value decomposition under the CTDE paradigm~\cite{liu2023CIA}. To enable effective CTDE, the Individual-Global-Max~(IGM) principle guarantees the optimal consistency between the joint and individual actions~\cite{QTRAN}. Based on this IGM principle, VDN~\cite{VDN} factorizes the joint action-value function into a summation of individual agent terms. In contrast, QMIX~\cite{QMIX} extends this additive value function factorization and imposes a monotonicity constraint. 
Furthermore, QTRAN~\cite{QTRAN}, WQMIX~\cite{WQMIX}, and QPLEX~\cite{QPLEX} progressively enlarged the family of functions that the mixing network can represent, while other advanced works exploit the value decomposition from the perspective of role-based learning~\cite{ROMA,RODE}, large-scale transfer learning~\cite{DBLP:conf/iclr/LongZ0FWW20,DyMA,DBLP:journals/corr/abs-2203-06416}, and intrinsic motivation-driven exploration~\cite{MAVEN,DBLP:conf/iclr/0001WWZ20,DBLP:conf/icml/JiangL21,DBLP:conf/nips/LiWWZYZ21,EMC}. However, these works still treat all entities as a whole and ignore their internal interaction information.

\subsection{Policy-Based MARL}

Another line of MARL methods under the CTDE paradigm deploys policy gradient to learn local policies with a centralized critic~\cite{MADDPG,COMA}. Prior works in this domain often directly employ the value decomposition mechanism to realize credit assignment in the centralized critic~\cite{LICA,VDAC,DOP,FACMAC}. Most of them follow the same off-policy fashion as value-based methods to avoid low sample efficiency. Recently, MAPPO~\cite{MAPPO} and HATRPO~\cite{HATRPO} impose the trust region constraint on multi-agent policy learning, and empirically show that on-policy policy-based methods can also provide competent efficiency and performance guarantees on many cooperative MARL benchmarks~\cite{papoudakis2020benchmarking,DBLP:conf/icml/FuYXYW22}. To further improve MAPPO, some advanced works attempt to alleviate the high variance issue in multi-agent policy gradient estimation~\cite{DBLP:conf/nips/WuYYZPZ21,DBLP:conf/nips/KubaWMGZMWY21,DBLP:conf/icml/LiXL22}. However, these works also encounter the difficulty of over-fitting~on~noisy~interactions.

\subsection{Model-Based MARL}

Model-based MARL aims to build the environment model to promote policy searching~\cite{wang2022model}. The multi-agent environment model often consists of not only a dynamics model but also multiple opponent models~\cite{DBLP:conf/ijcai/0001WSZ21}. Sessa et al.~\cite{DBLP:conf/icml/SessaK022} propose to construct confidence intervals around the transition function and hallucinate optimistic value functions for agents to guide exploration.
Zhang et al.~\cite{zhang2020model} and Liu et al.~\cite{liu2021sharp} attempt to establish the theoretical analysis of sample complexity of model-based MARL in zero-sum Markov games. 
Moreover, Mahajan et al.~\cite{Tesseract} and Van Der Vaart et al.~\cite{van2021model} factorize the action-value function based on tensor decomposition to model agent interactions, and further develop an effective model-based algorithm using low-rank transition and reward functions. This tensor decomposition mechanism, however, is not applicable to our setting where non-agent entities should also be considered.

\subsection{Interaction Pattern in MARL}

Exploring complex interaction patterns has been widely studied in MARL. Several works~\cite{DGN,NCC,DBLP:conf/aaai/RyuSP20} capture multi-agent interplay using the pre-defined graph. However, these prior relationships between entities are often unavailable. On the other hand, MAAC~\cite{MAAC} and ATOC~\cite{ATOC} introduce the attention mechanism to learn the interaction relationships, while UPDeT~\cite{UPDeT} and PIT~\cite{zhou2021cooperative} utilize attention-based semantic alignment between the input entities and the output actions to decompose the policy. Despite the promising results achieved, these works mainly rely on dense attention and have to focus on irrelevant entities. To remedy this issue, ALC-MADDPG~\cite{DBLP:conf/aaai/ZhangYAZ21} and CASEC~\cite{CASEC} use the predefined threshold to prune the low correlation relationships between agents. Moreover, G2ANet~\cite{G2ANet} and S2RL~\cite{luo2022s2rl} introduce the sparse attention mechanism to learn the dynamics among agents. 
Mahajan et al.~\cite{mahajan2022generalization} formalize the multi-agent generalization problem and provide theoretical bounds on generalization.
Furthermore, REFIL~\cite{REFIL} randomly partitions entities into sparse subgroups and forces each subgroup to learn the same objective function as all entities. This random grouping strategy serves as a regularization term to improve the generalization ability of value functions.
Nevertheless, the interaction patterns in these methods still remain entangled. They only consider the sparsity of interaction patterns but ignore the diversity. Different interaction patterns often simultaneously exist among entities in a complex task. To this end, we introduce a novel method to realize the interaction pattern disentangling for both sparsity and diversity.

\section{Preliminary\label{sec:background}}

In this section, we formally define the cooperative MARL problem under the \textit{decentralized partially observable Markov decision process}~(\text{Dec-POMDP}) with entities. Then we introduce the \textit{centralized training and decentralized execution}~(CTDE) paradigm for value decomposition.

\subsection{Dec-POMDP with Entities}
Cooperative MARL problem can be described as a \text{Dec-POMDP} with entities~\cite{Dec-POMDP,REFIL} defined by a tuple $\langle\mathcal{E},\mathcal{A},\mathcal{T},\mathcal{S},\mathcal{U},P,r,\Omega,O,\gamma \rangle$, where $\mathcal{E}$ is the set of entities in the environment. Each entity $e$ has a state representation $s^e$ and $s = \{s^e \mid e\in \mathcal{E} \} \in \mathcal{S}$ is the global state of the environment. $\mathcal{A} \subseteq \mathcal{E}$ is the finite set of agents, while the non-agent entities are part of the environment (e.g. uncontrollable enemies, obstacles and landmarks). $\mathcal{T}$ is the set of tasks, where each task is composed of a different number of entities and agents. At each time step $t$, each agent $a \in \mathcal{A}$ receives an individual partial observation $o^a_t \in \Omega$ according to the observation function $O(s_t, a)$ and chooses an action $u^a_t \in \mathcal{U}$, forming a joint action $\boldsymbol{u}_t$. This causes a transition to the next state $s_{t+1}$ according to the state transition function $P(s_{t+1} | s_t, \boldsymbol{u}_t)$. All agents share the same reward function $r(s_t, \boldsymbol{u}_t)$ and $\gamma \in [0, 1)$ is the discount factor. Each agent $i$ has an action-observation history $\tau^a \in \mathcal{T} \equiv (\Omega \times \mathcal{U})^*$ and learns its individual policy $\pi^a(u^a|\tau^a)$ to jointly maximize the discounted return $R_t = \sum_{k=0}^{\infty}{\gamma^k r_{t+k}}$. $\boldsymbol{\tau}$ is used to denote joint action-observation history. The joint policy $\boldsymbol{\pi}$ induces a joint action-value function $Q^{\boldsymbol{\pi}}_{tot}(s_t,\boldsymbol{u}_t)=\mathbb{E}_{s_{t+1:\infty},\boldsymbol{u}_{t+1:\infty}}{\left[R_t \mid s_t,\boldsymbol{u}_t\right]}$. 

\subsection{CTDE Paradigm}
The CTDE paradigm allows agents to learn their individual utility functions by optimizing the joint action-value function for credit assignment~\cite{RES,FOP,QPD,ASN}. During centralized training, the learning model has access to the global state and the action-observation histories of all agents, whereas each agent can only condition on its own action-observation history during decentralized execution. Due to the partial observability, $Q_{tot}(\boldsymbol{\tau},\boldsymbol{u},s)$ is used in place of $Q_{tot}(\boldsymbol{u},s)$. A mixing network is introduced to merge all individual action values into a joint one $Q_{tot}(\boldsymbol{\tau},\boldsymbol{u},s;\theta_{\upsilon }) = f([Q_a(\tau^a,u^a;\theta_{\pi})]_{a\in\mathcal{A}},s;\theta_{\upsilon })$, where $Q_a$ is the utility network of each agent $a$. The learnable parameter $\theta=\{\theta_{\pi},\theta_{\upsilon }\}$ can be updated by minimizing the following expected Temporal-Difference~(TD) loss:
\begin{align}
    \mathcal{L}_{TD}(\theta) =  \mathbb{E}_{\mathcal{D}} \left[\left(y^{tot} - Q_{tot}(\boldsymbol{\tau},\boldsymbol{u},s;\theta_{\upsilon })\right)^2\right].
\end{align}
where $\mathcal{D}$ is the replay buffer of the transitions, $y^{tot}=r+\gamma\max_{\boldsymbol{u}'}Q_{tot}(\boldsymbol{\tau}',\boldsymbol{u}',s';\theta^-_{\upsilon })$ is the target value and $\theta^-_{\upsilon }$ is the parameter of the target network~\cite{DQN15}.

\section{Method\label{sec:method}}

In what follows, we first provide an overview of the framework based on the proposed \emph{interactiOn Pattern disenTangling}~(OPT) module. 
Then we further detail the OPT module and summarize the complete optimization objective.

\begin{figure}[!t]
  \centering
  \includegraphics[width=0.48\textwidth]{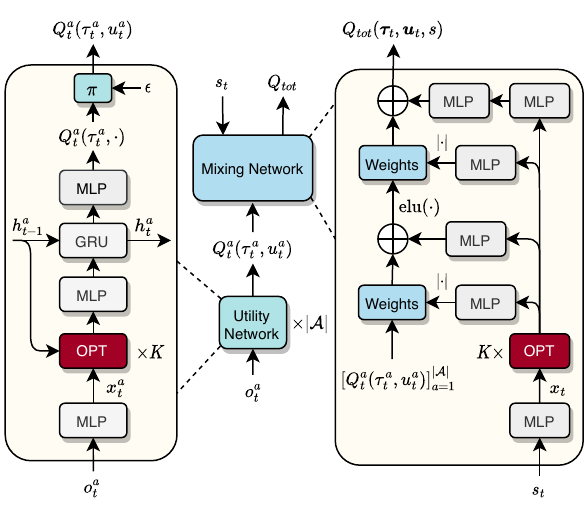}
  \caption{An overview of the framework based on the proposed \emph{interactiOn Pattern disenTangling}~(OPT) method. The middle is the basic MARL framework under the CTDE paradigm, where we use the OPT module in both the utility network and mixing network. The left is the utility network of each agent, and the right is the mixing network. }
  \label{fig:process}
\end{figure}

\subsection{Overall Framework}

Our framework adopts the CTDE paradigm, where each agent learns its individual utility network by optimizing the TD loss of the mixing network. 
As shown in Figure~\ref{fig:process}, the OPT serves as a plug-in module in both the utility network and mixing network, jointly facilitating the learning of decision making and credit assignment. 
We use a Transformer architecture~\cite{DBLP:conf/nips/VaswaniSPUJGKP17} to stack multiple layers of OPT modules, which can promote the generalization ability of the model. The outputs of the OPT modules are directly used for subsequent network computations. All the network modules, including OPT, MLP and GRU, perform entity-wise operations on their input.

Specially, the utility network of each agent $a$ takes the current observation $o_t^a$ and the history information $h_{t-1}^a$ as input, estimating the individual action value $Q^a_t(\tau_t^a,\,\cdot\,)$ based on the OPT module. Moreover, each action value is calculated by its corresponding entity~\cite{UPDeT,REFIL}. Then the agent $a$ selects an action $u_t^a$ that maximizes the action value $Q^a_t(\tau_t^a,\,\cdot\,)$ to execute in the environment, while the action value $Q^a_t(\tau_t^a,u_t^a)$ is used as the input to the mixing network. Here we adopt the annealed $\epsilon$-greedy strategy for exploration. 
Our mixing network implementation uses QMIX~\cite{QMIX} as a basic backbone for its robust performance, but it is readily applicable to the other mixing methods. The mixing network is a feed-forward neural network that merges all individual action values into a joint one $Q_{tot}(\boldsymbol{\tau},\boldsymbol{u},s;\theta_{\upsilon })$, where the weights of the mixing network are generated by a hyper-network that takes the global state $s_t$ as input. The OPT module is used to handle the global state $s_t$ in this hyper-network of the mixing network.

\begin{figure*}[!t]
  \centering
  \includegraphics[scale=0.92]{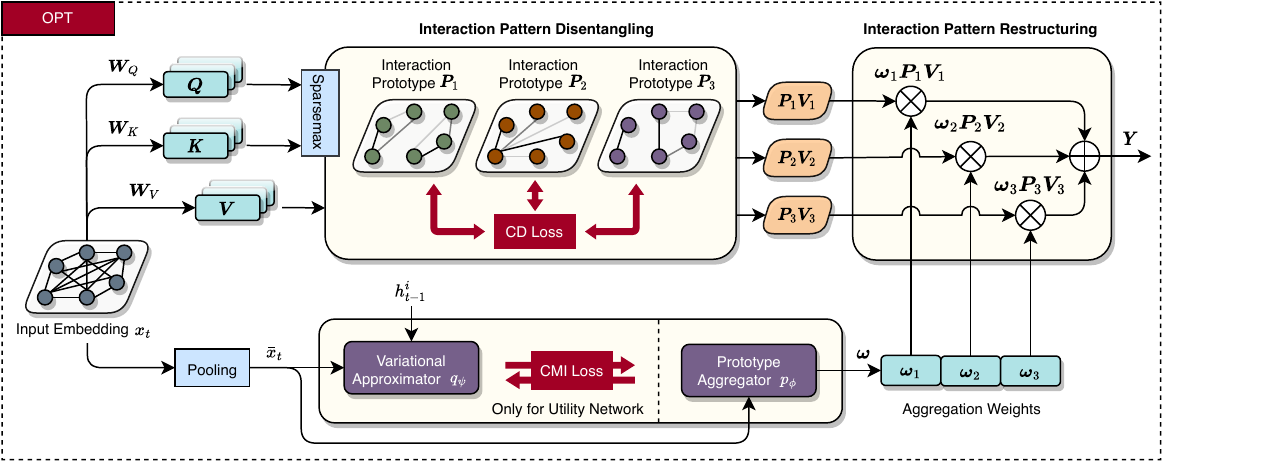}
  \caption{An illustrative diagram of the proposed \emph{interactiOn Pattern disenTangling} (OPT) method, which mainly consists of two steps: disentangling and restructuring.}
  \label{fig:method}
\end{figure*}

Intuitively, we can explain the intrinsic mechanism of OPT from the perspective of subgraph mining. The disentangled interaction prototypes represent several interaction subgraphs. The method we adapt to generate these prototypes is based on the self-attention module in Transformer, which can capture the relationship weights between each pair of entities. 
As shown in Figure~\ref{fig:method}, we start by introducing the explicit disentangling step to encourage sparsity and diversity among discovered prototypes. Then we selectively restructure these prototypes by a learnable aggregator, unweighting the irrelevant prototypes and emphasizing the important ones.

\subsection{Disentangling}

The goal of the disentangling step is to explore the diverse interaction prototypes from the input observation or state of entities. 
The disentangling steps of the utility network and mixing network are similar except for different inputs. Thus, we use the mixing network as an example to describe this process.
At each time step $t$, the input state of the mixing network is defined as $s_t = [s_t^1,\dots,s_t^M]^T \in \mathbb{R}^{M \times d_e}$, where $M$ is the number of the entities and $d_e$ is the state representation dimension of the entity. We first encode the state representation of each entity $s_t^e$ via an entity-wise embedding function $f(\cdot):\mathbb{R}^{d_e}\to\mathbb{R}^{d_x}$ and output the embedding feature $x_t^e = f(s_t^e)$ with the dimension of $d_x$. Then the input of the disentangling step is denoted as $x_t = [x_t^1,\dots,x_t^M]^T \in \mathbb{R}^{M \times d_x}$. We assume that entities treat each other equally before disentangling, forming a fully connected interaction pattern. Therefore, our method focuses on disentangling this fully connected pattern into several interaction prototypes.

Each interaction prototype can be described as an attention map where the relationship weights between entities are computed by the query with the corresponding key. To simplify the following description, the embedding matrix $\textbf{X}$ is used instead of $x_t$. Specifically, we first transform the input embedding $\textbf{X}$ into queries $\textbf{Q} \in \mathbb{R}^{M \times d_x}$, keys $\textbf{K} \in \mathbb{R}^{M \times d_x}$, and values $\textbf{V} \in \mathbb{R}^{M \times d_x}$:
\begin{align}
    \begin{bmatrix} \textbf{Q} \\ \textbf{K} \\ \textbf{V} \end{bmatrix} = \textbf{X} \begin{bmatrix} \textbf{W}_Q \\ \textbf{W}_K \\ \textbf{W}_V \end{bmatrix}, 
\end{align}
where $\varphi = \{\textbf{W}_Q, \textbf{W}_K, \textbf{W}_V\}$ are parameter matrices and each matrix $ \textbf{W} \in \mathbb{R}^{d_x \times d_x}$ is trainable. Then we can calculate the dot product for queries and keys to obtain the relationship weights between entities, followed by an activation function to normalize the output weights to a probability distribution.

The conventional activation function used for normalization is the softmax function, which never assigns a probability of zero to any weights. Due to the strictly positive probabilities, we can never rule out the irrelevant entities for the interaction prototypes. This density is wasteful and noisy, making the exploration space more complex. In a particular interaction prototype, an entity usually interacts with only several crucial entities, which implies a sparsity constraint over the probability distribution. Thus, a sparse activation function is desired to focus on the important entities and ignore the irrelevant entities. Inspired by the sparse mechanism used in \cite{sparsemax,DBLP:conf/emnlp/CorreiaNM19}, we introduce sparsemax as an alternative to softmax, which tends to yield sparse probability distribution: 
\begin{align}
    \operatorname{sparsemax}(\boldsymbol{z}) := \mathop{\arg\min}_{\boldsymbol{p} \in \Delta^M} \Vert \boldsymbol{p} - \boldsymbol{z} \Vert_2,
    \label{eq:sparse}
\end{align}
where $\boldsymbol{z} \in \mathbb{R}^M$ is the input vector and $\boldsymbol{p}\in \mathbb{R}^M$ is the output vector. $\Delta^M := \big\{\boldsymbol{p} \in \mathbb{R}^M \mid \boldsymbol{p} \geq 0, \Vert\boldsymbol{p}\Vert_1=1\big\}$ is the probability simplex. Sparsemax realizes a Euclidean projection of the input vector onto the probability simplex, which is likely to hit the boundary of the simplex and obtain the sparse solution. The solution of Eq.~(\ref{eq:sparse}) is obtained as follows:
\begin{align}
  \boldsymbol{p} = [\boldsymbol{z} - \sigma(\boldsymbol{z})\boldsymbol{1}]_+,
\end{align}
where $[\cdot]_+ = \max(0, \cdot)$ is a clipping function and $\boldsymbol{1}\in\mathbb{R}^{M}$ denotes the all-one vector. $\sigma(\cdot): \mathbb{R}^{M}\to\mathbb{R}$ is a threshold function. We first sort the elements of vector $\boldsymbol{z}$ in descending order to derive a sorted vector $\boldsymbol{\hat{z}}$. Then the number of non-zero elements in the sparse solution $\boldsymbol{p}$ is calculated as 
\begin{align}
m(\boldsymbol{\hat{z}}) = \mathop{\arg\max}_{m\in \{1,2,\cdots,M\}} \left\{ m\boldsymbol{\hat{z}}_{m} > \sum_{i\leq m}\boldsymbol{\hat{z}}_{i} - 1 \right\}. 
\end{align}
Thus, the threshold value is given by 
\begin{align}
  \sigma(\boldsymbol{z}) =  \frac{\left( \sum_{i\leq m(\boldsymbol{\hat{z}})}\boldsymbol{\hat{z}}_{i} - 1 \right)}{m(\boldsymbol{\hat{z}})}.
\end{align}
It is easy to prove that $\sigma(\boldsymbol{z}) $ satisfies $\sum_{i}\left[\boldsymbol{z}_i - \sigma(\boldsymbol{z})\right]_+ = 1$ for every $\boldsymbol{z}$.
In this way, we can directly calculate the sparse probability distribution based on the threshold value. Sparsemax can retain most of the important properties of softmax, and meanwhile assign exactly zero probability to low-scoring choices.

This sparsity enables us to discover similar subgroup entities and promotes the generalization of the generated interaction prototypes. Consequently, an interaction prototype is constructed by
\begin{align}
    \textbf{P} = \operatorname{sparsemax}\left(\frac{\textbf{Q}\textbf{K}^T}{\sqrt{d_x}}\right),
\end{align}
where $\sqrt{d_x}$ is the scaling coefficient. We construct $N$ interaction prototypes. Each prototype can be represented by $\textbf{P}_n \in \mathbb{R}^{M\times M}$ with its own trainable parameter matrices $\varphi_n$. Moreover, the corresponding value matrix is defined as $\textbf{V}_n$. Then these prototypes will be used for the next restructuring step. However, without any other constraint, some generated interaction prototypes may contain a similar structure, degrading the disentanglement performance and capacity of the model. We therefore design an additional loss function in the disentangle step, aiming to avoid the similarity of the generated interaction prototypes.

The motivation of the additional loss function is that, a well disentangled interaction prototype should be clearly distinguished from the others. However, obtaining the solution that all the disentangled interaction prototypes differ from each other without any constraint will severely damage the performance. The distance between different prototypes can be infinitely enlarged, resulting in divergence of the optimization process. We thus constrain the solution by using a contrastive disagreement~(CD) loss:
\begin{align}
    \mathcal{L}_{CD}(\theta,\varphi) = \mathbb{E}_{n,e} \left[ -\log\left( \frac{\exp{\left(\left(\textbf{P}_n^e \textbf{V}_n\right)^T\textbf{P}_n^e\textbf{V}_n\right)}}{\sum_{i=1}^{N}{\exp{\left(\left(\textbf{P}_n^e\textbf{V}_n\right)^T\textbf{P}_i^e\textbf{V}_i\right)}}}  \right)  \right],
\end{align}
where $\textbf{P}_n^e \in \mathbb{R}^{1\times M}$ is the interaction prototype of the entity $e$. In the contrastive disagreement loss, we use the prototype itself as the positive sample and the other prototypes as the negative sample. In this way, this loss only focuses on the negative prototypes and constrains them to uniformly distribute on a hypersphere without divergence~\cite{wang2020understanding}. With the contrastive disagreement loss, we directly encourage the diversity among interaction prototypes. This diversity enables the model to provide discriminative interaction prototypes for different tasks.

\subsection{Restructuring}
In the restructuring step, we aim to selectively restructure the interaction prototypes to form the final compact pattern. The input embedding $x_t \in \mathbb{R}^{M \times d_x}$ of the restructuring step is the same as the previous disentangling step. 
Firstly, we use a mean pooling operation to extract the global information $\bar{x}_t \in \mathbb{R}^{d_x}$ from $x_t$ over all entities. 
Then the prototype aggregator $p_{\phi}: \mathbb{R}^{d_x} \to \mathbb{R}^{N}$ takes $\bar{x}_t$ as input to derive the aggregation weights $\boldsymbol{\omega} \in \mathbb{R}^{N}$. 
The prototype aggregator $p_{\phi}$ parameterized by ${\phi}$ is a one-layer MLP network with softmax function. Then we use the aggregation weights $\boldsymbol{\omega}$ to restructure the interaction prototypes and output the final compact pattern $\textbf{Y} = \sum_{n=1}^{N} \boldsymbol{\omega}_n \textbf{P}_n\textbf{V}_n \in \mathbb{R}^{M \times d_x}$.
The restructuring processing could facilitate the extraction of global information while adapting to different interaction prototypes with learnable weights. Then the output pattern is employed for the following decision making in the utility network or credit assignment in the mixing network. 

Although the global state could be computed directly in the mixing network, the observation information in the utility network is incomplete due to the partial observability, resulting in the instability of the aggregation weights. To prevent this, we propose a conditional mutual information~(CMI) objective:
\begin{align}
    I(\boldsymbol{\omega}^a_t ; \tau^a_{t-1} | o^a_t) = H(\boldsymbol{\omega}^a_t | o^a_t) - H(\boldsymbol{\omega}^a_t | \tau^a_{t-1}, o^a_t)
\end{align}
where $H$ is the entropy and the aggregator of the agent $a$ is regularized not only conditioned on its observation but also its history behaviors. Therefore, we can stabilize the generated aggregation weights over time via maximizing the CMI objective. 
However, directly optimizing this objective is quite difficult, so we design a variational approximator $q_{\psi}$ parameterized by $\psi$ to approximate the posterior over $\boldsymbol{\omega}^a_t$ given the observed trajectories~\cite{ROMA,MAVEN}.
The variational inference provides a lower bound of the CMI objective to optimize, which can be formalized as a loss function:
\begin{align}
    \mathcal{L}_{CMI}(\theta,\phi,\psi) = \mathbb{E}_{\mathcal{D}} \Bigl[ \operatorname{KL}\bigl[p_{\phi }(\boldsymbol{\omega}^a_t | \bar{x}_t^a) \parallel q_{\psi }(\boldsymbol{\omega}^a_t | h^a_{t-1}, \bar{x}_t^a)\bigl]\Bigl],
\end{align}
where $\operatorname{KL}(\cdot)$ is the KL divergence function, $\bar{x}_t^a$ is the input embedding of the observation $o^a_t$ after mean pooling operation, and a GRU network~\cite{GRU} is used to encode the history trajectory $\tau^a_{t-1}$ into $h^a_{t-1}$. Compared to directly relying on the aggregator on history behaviors, the CMI objective derives a tractable lower bound, which provides a more solid objective to optimize. The proof is given in the appendix.

\subsection{Optimization Objective}
To sum up, training our framework based on the OPT module contains three main parts. The first one is the original TD loss $\mathcal{L}_{TD}$, which enables each agent to learn its individual policy by optimizing the joint action value of the mixing network. The second one is the CD loss $\mathcal{L}_{CD}$ to facilitate the sparsity and diversity among interaction prototypes when disentangling. The last one is the CMI loss $\mathcal{L}_{CMI}$ to stabilize the aggregation weights for restructuring.
Thus, given the three corresponding loss items, the total loss of our framework is formulated as follows:
\begin{align}
    \mathcal{L}(\theta,\varphi,\phi,\psi) = \mathcal{L}_{TD}(\theta) + \alpha  \mathcal{L}_{CD}(\theta,\varphi) + \beta  \mathcal{L}_{CMI}(\theta,\phi,\psi),
\end{align}
where the $\alpha $ and $\beta$ are coefficients. The overall framework is trained in an end-to-end centralized manner. We summarize the full procedure in Algorithm~\ref{alg:alg}.

\textbf{Complexity Analysis.} For simplicity, the dimensions of all entity features and hidden features are assumed $d$. $M$ is the number of the entities. The operations of all network modules are entity-wise. The OPT module consists of feed-forward and sparse-attention networks with the time complexity of $\mathcal{O}(Md^2+M^2d)$. 
Thus, the time complexity of the overall framework is $\mathcal{O}\left(|\mathcal{A}|(Md^2+M^2d) + (|\mathcal{A}|d + Md^2+M^2d) \right)$, where the former refers to the utility network process of $|\mathcal{A}|$ agents and the latter refers to the mixing network process.

\algdef{SE}[SUBALG]{Indent}{EndIndent}{}{\algorithmicend\ }%
\algtext*{Indent}
\algtext*{EndIndent}
\algnewcommand{\LineComment}[1]{\State \textcolor{gray}{\(\triangleright\) #1}}

\begin{algorithm}[!t]
  \caption{OPT for Multi-Agent Reinforcement Learning}
  \label{alg:alg}
  \begin{flushleft}
    \textbf{Initialize:} Utility network $\theta_{\pi}$, mixing network $\theta_{\upsilon}$, target network $\theta_{\upsilon}^-=\theta_{\upsilon}$, replay buffer $\mathcal{D}$ \\
  \end{flushleft}
  \begin{algorithmic}[1]
  \Repeat
    \LineComment{Collect the data}
    \State Sample a task from $\mathcal{T}$
    \State Obtain the initial global state $s_0$
    \While{not terminal}
      \For{each agent $a$}:
        \State Obtain the observation $o_t^a = O(s_t,a)$
        \State Calculate the input embedding $x_t^a$
        \LineComment{Disentangling}
        \State Calculate $\{\textbf{Q}_{t,n}^a, \textbf{K}_{t,n}^a, \textbf{V}_{t,n}^a\}_{n=1}^N$
        \State Extract the interaction prototypes $\{\textbf{P}_{t,n}^a\}_{n=1}^N$
        \LineComment{Restructuring}
        \State Calculate the aggregation weights $\boldsymbol{\omega}^a_t$
        \State Construct the final compact patterns $\textbf{Y}_{t,n}^a$
        \LineComment{Policy}
        \State Calculate $h_{t}^a = \operatorname{GRU}(\textbf{Y}_{t,n}^a, h_{t-1}^a)$
        \State Calculate the action values $Q_t^a(h_{t}^a, \cdot)$
        \State Sample $u_t^a$ from $Q_t^a(h_{t}^a, \cdot)$ based on $\epsilon$-greedy
      \EndFor
      \State Execute the joint action $\boldsymbol{u}_t = [u_t^1,\cdots,u_t^{|\mathcal{A}|}]$
      \State Receive the reward $r_{t+1}$ and the next state $s_{t+1}$
    \EndWhile
    \State Store the episode to the replay buffer $\mathcal{D}$
    \LineComment{Train the networks}
    \State Sample episodes from the replay buffer $\mathcal{D}$
    \State Calculate joint action values with mixing network
    \State Compute $\mathcal{L}_{TD}(\theta)$, $\mathcal{L}_{CD}(\theta,\varphi)$,  $\mathcal{L}_{CMI}(\theta,\phi,\psi)$
    \State Update the network parameters $\theta,\varphi,\phi,\psi$
    \State Update the target network $\theta_{\upsilon}^- = \theta_{\upsilon}$ every $C$ episodes
  \Until converge
  \end{algorithmic}
\end{algorithm}

\begin{figure*}[!t]
  \centering
  \subfloat{\quad\quad\includegraphics[scale=0.75]{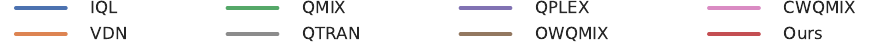}}\\    
    \addtocounter{subfigure}{-1}
    \vspace{-0.2cm}
  \subfloat[10m\_vs\_11m (Easy)]{\includegraphics[width=0.333\textwidth]{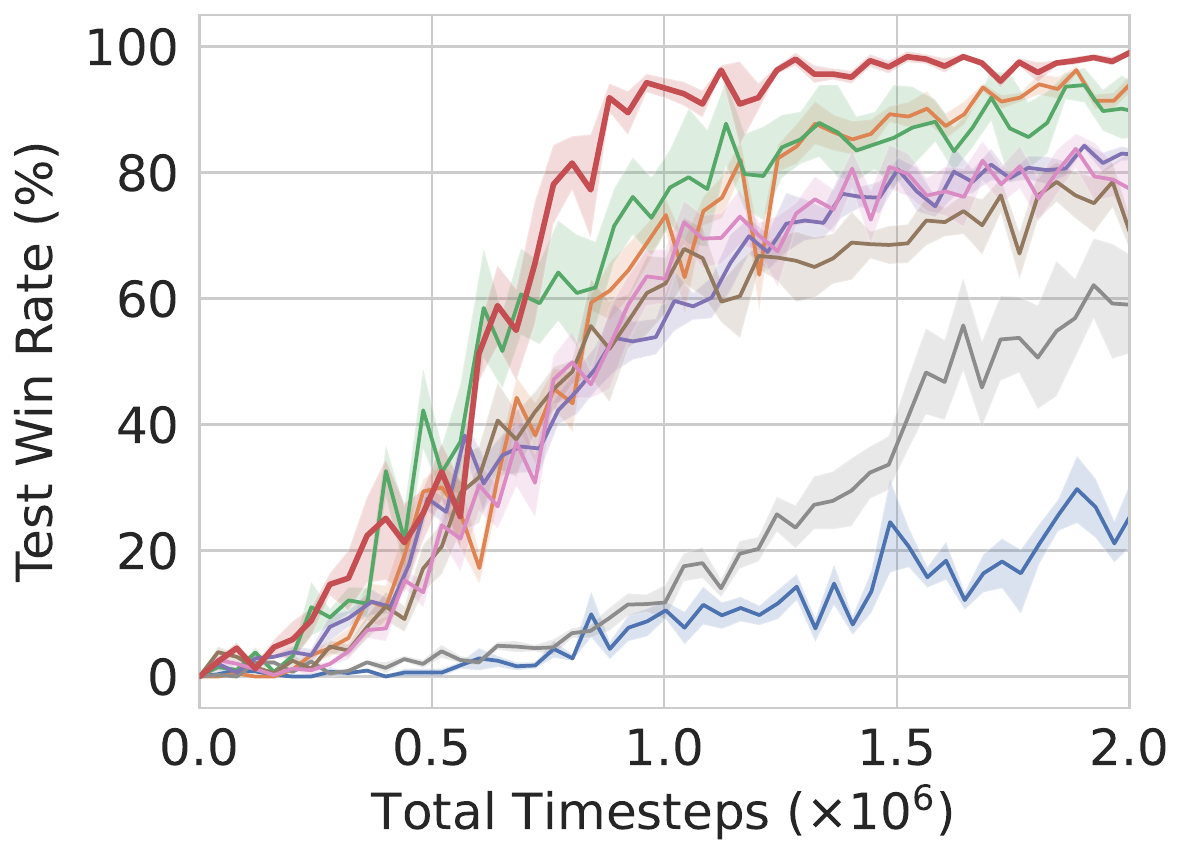}}\hfill
  \subfloat[5m\_vs\_6m (Hard)]{\includegraphics[width=0.333\textwidth]{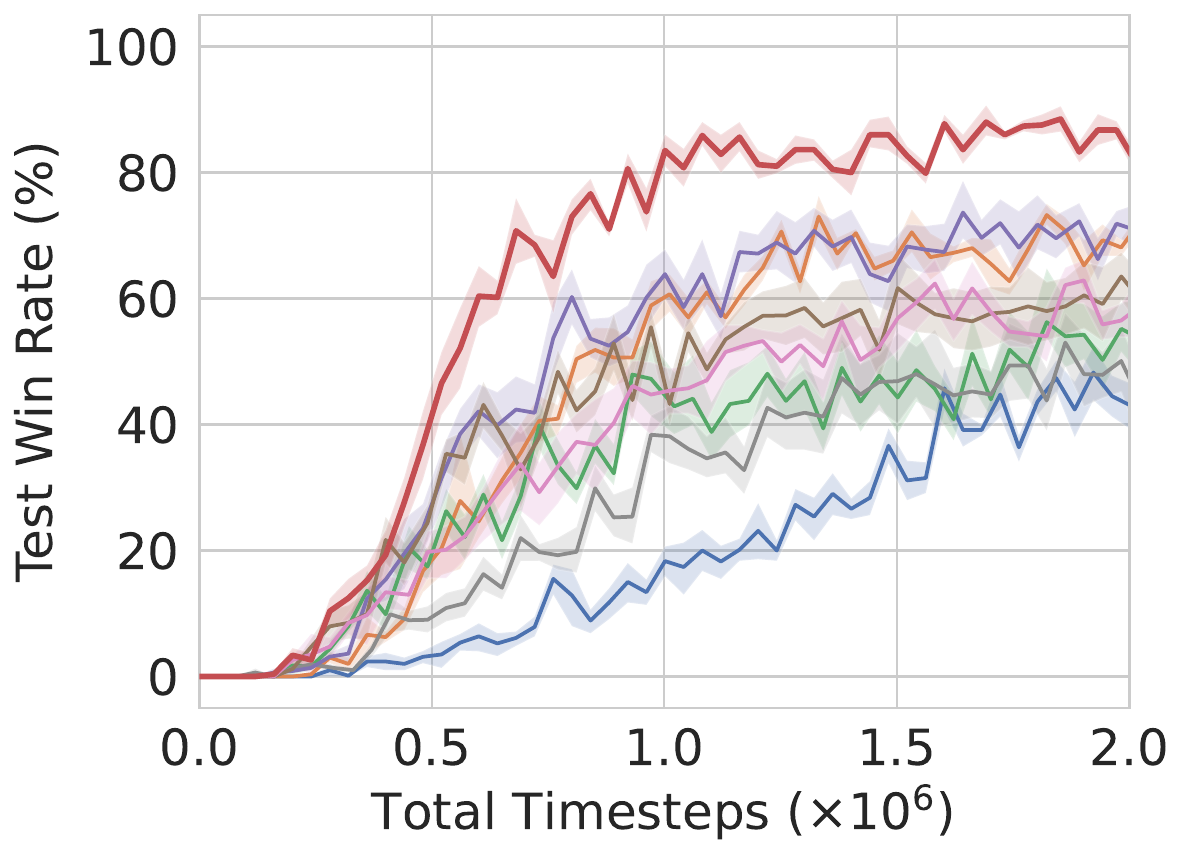}}\hfill
  \subfloat[MMM2 (Super Hard)]{\includegraphics[width=0.333\textwidth]{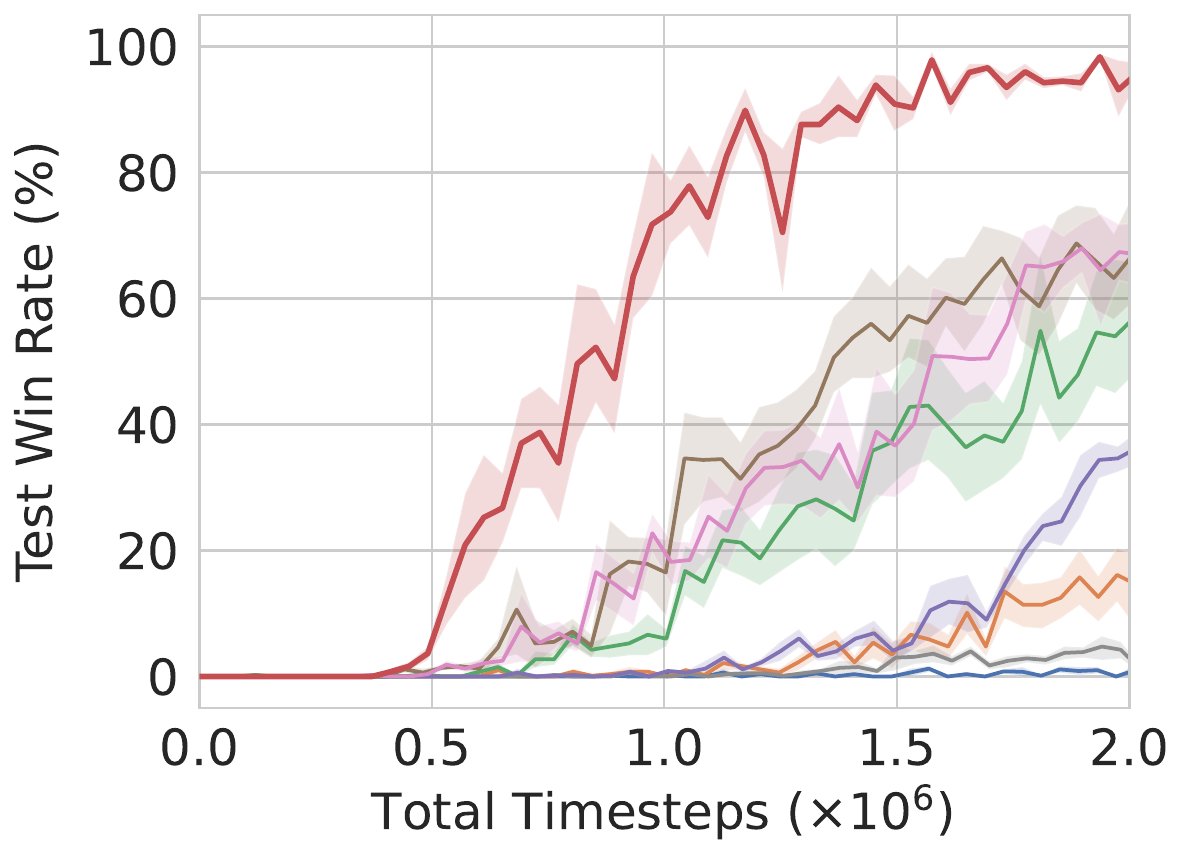}}

  \subfloat[corridor (Super Hard)]{\includegraphics[width=0.333\textwidth]{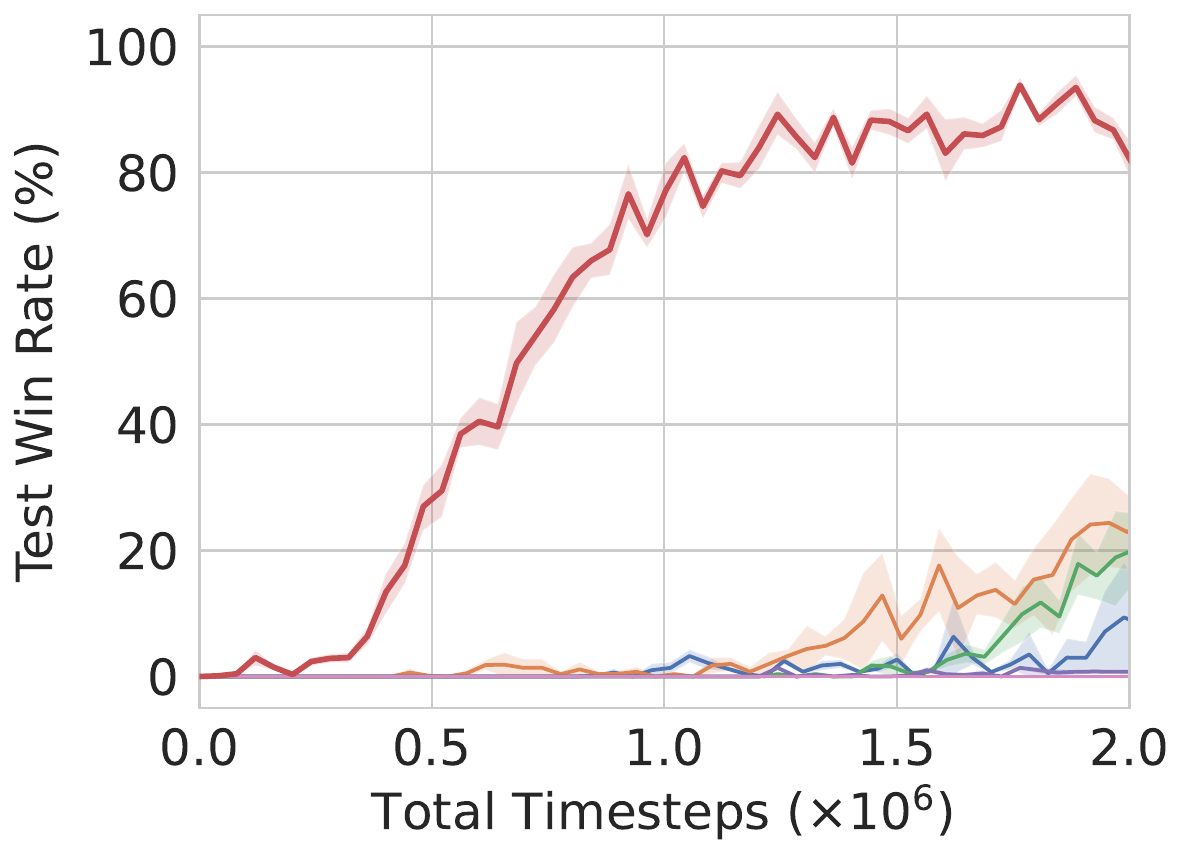}}\hfill
  \subfloat[6h\_vs\_8z (Super Hard)]{\includegraphics[width=0.333\textwidth]{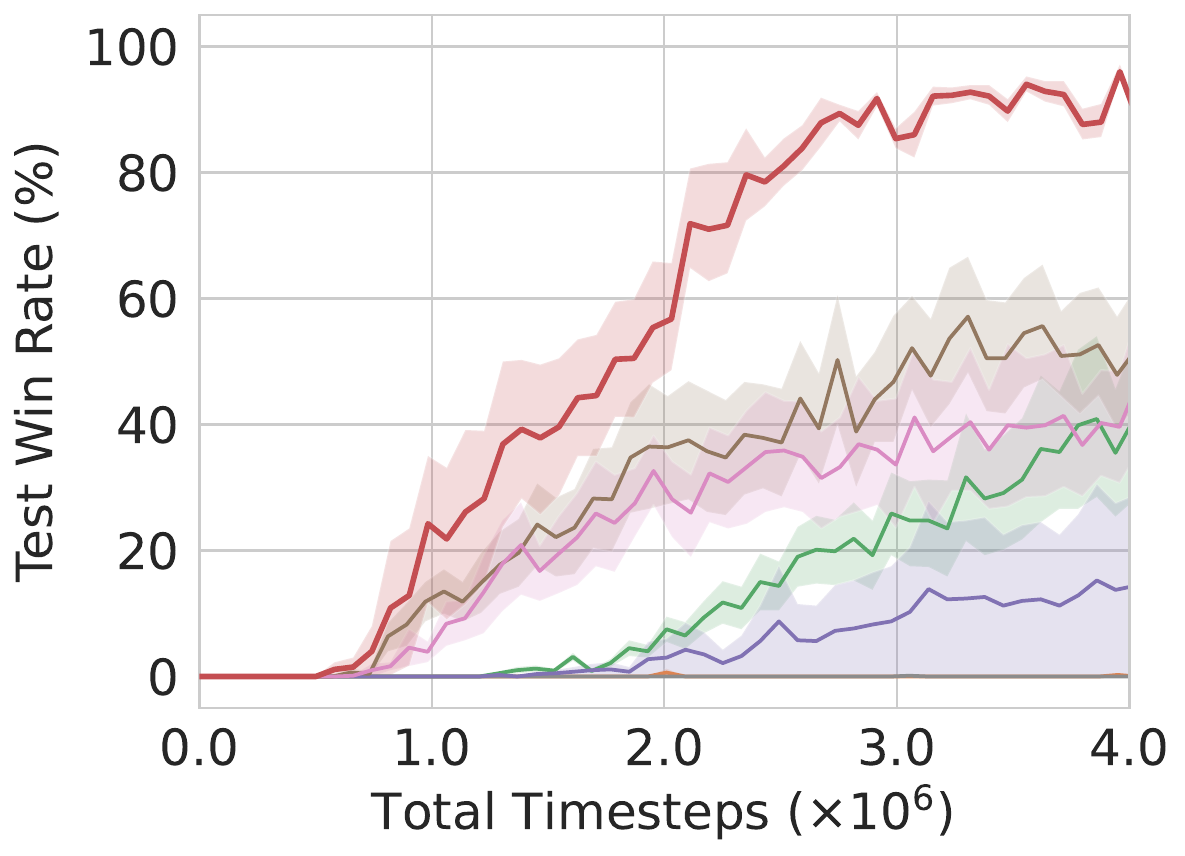}}\hfill
  \subfloat[3s5z\_vs\_3s6z (Super Hard)]{\includegraphics[width=0.333\textwidth]{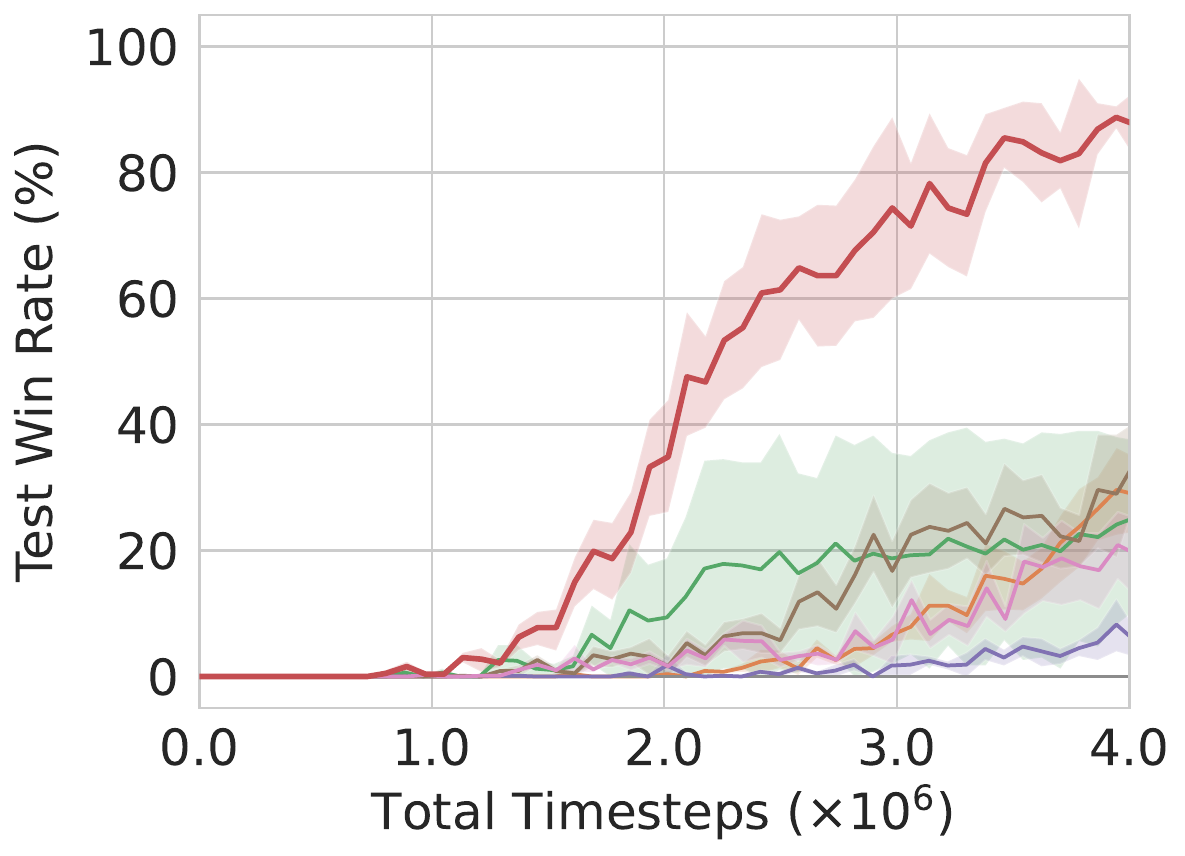}}
  \caption{Learning curves of our method and value-based baselines in 6 single-task SMAC scenarios. All experimental results are illustrated with the mean and the standard deviation of the performance over five random seeds for a fair comparison. To make the results in figures clearer for readers, we adopt a 50\% confidence interval to plot the error region.}
  \label{fig:sigle-task}
  \vspace{-0.05cm}
\end{figure*}

\begin{figure*}[!t]
  \centering
  \subfloat{\quad\quad\includegraphics[scale=0.8]{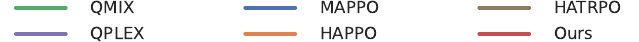}}\\    
    \addtocounter{subfigure}{-1}
    \vspace{-0.2cm}
  \subfloat[10m\_vs\_11m (Easy)]{\includegraphics[width=0.333\textwidth]{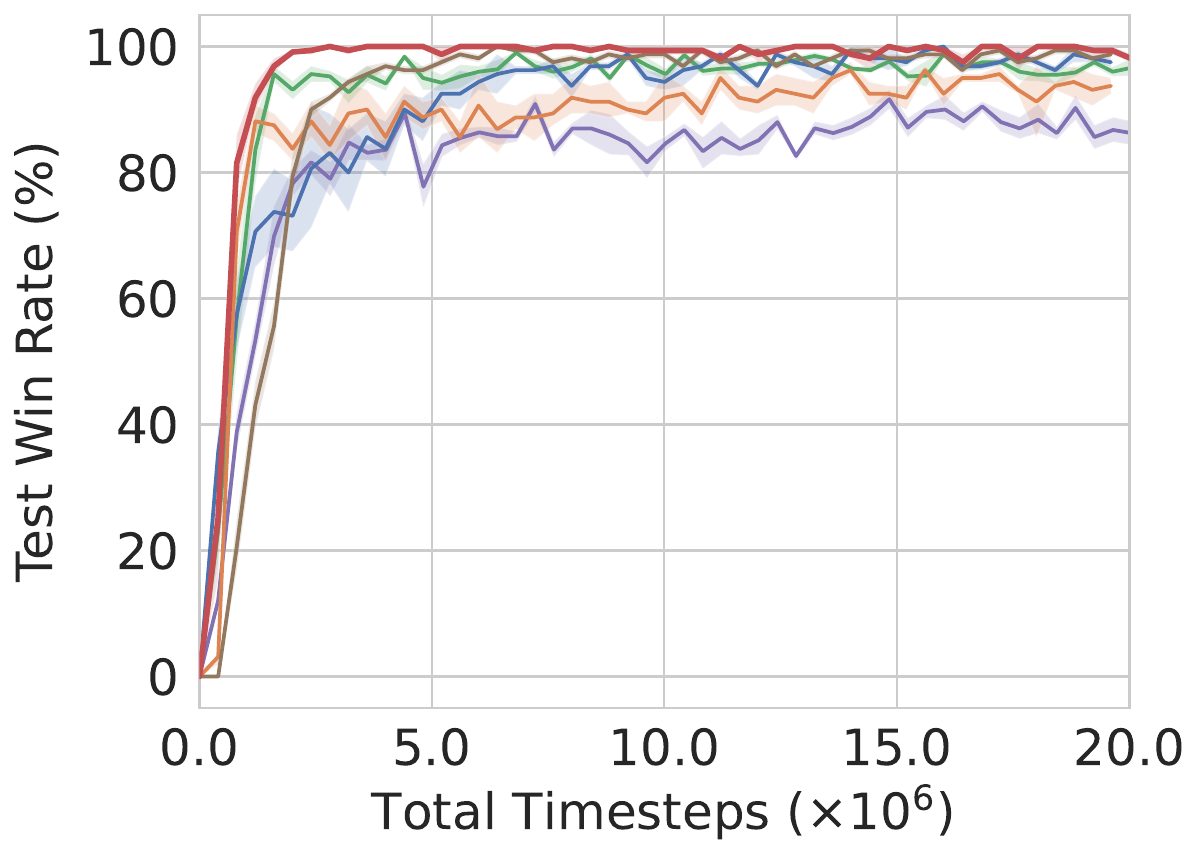}}\hfill
  \subfloat[5m\_vs\_6m (Hard)]{\includegraphics[width=0.333\textwidth]{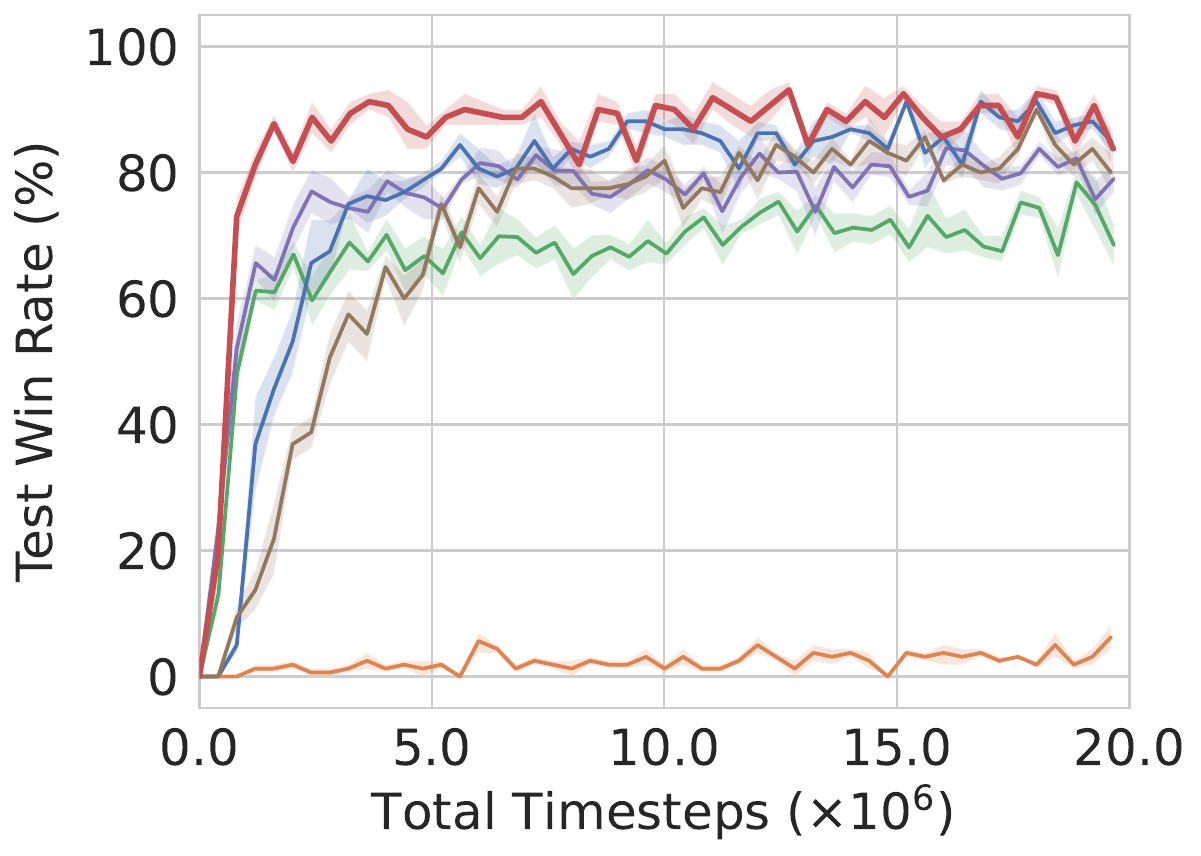}}\hfill
  \subfloat[MMM2 (Super Hard)]{\includegraphics[width=0.333\textwidth]{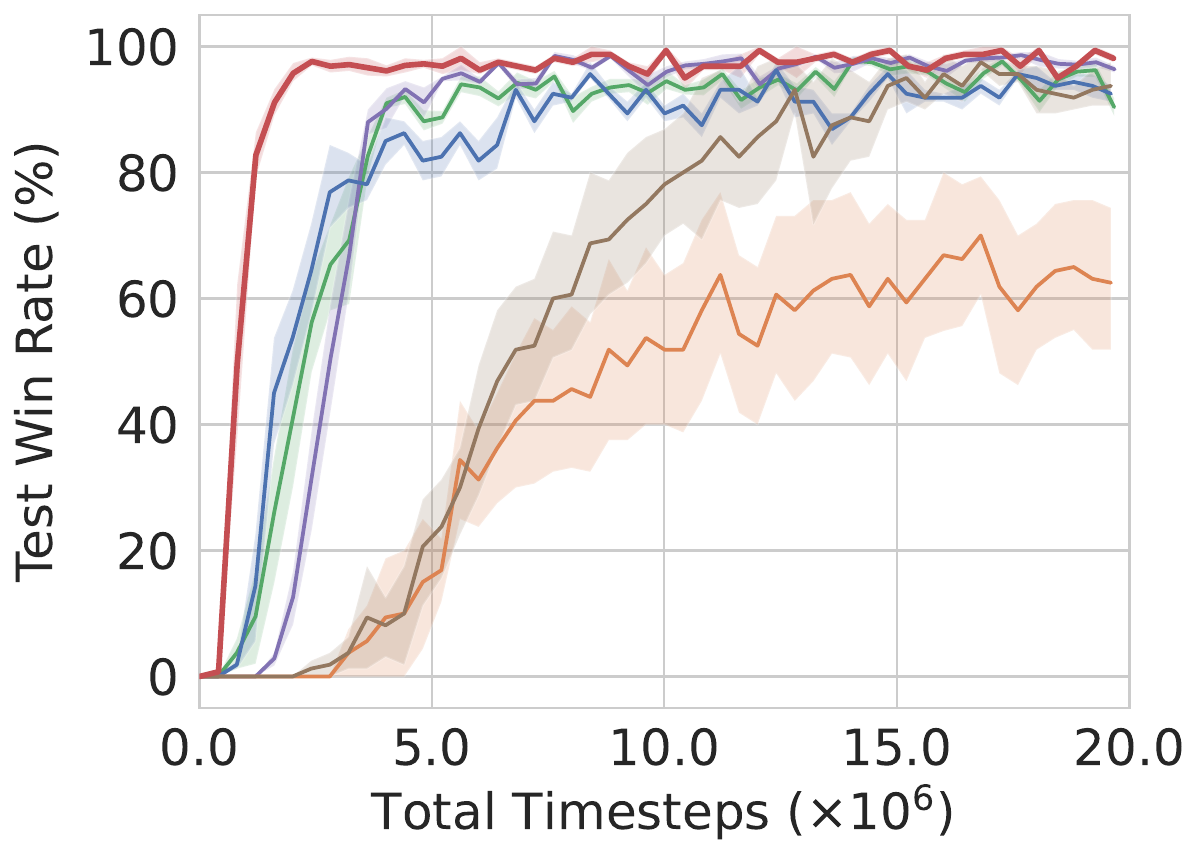}}\\
  \subfloat[corridor (Super Hard)]{\includegraphics[width=0.333\textwidth]{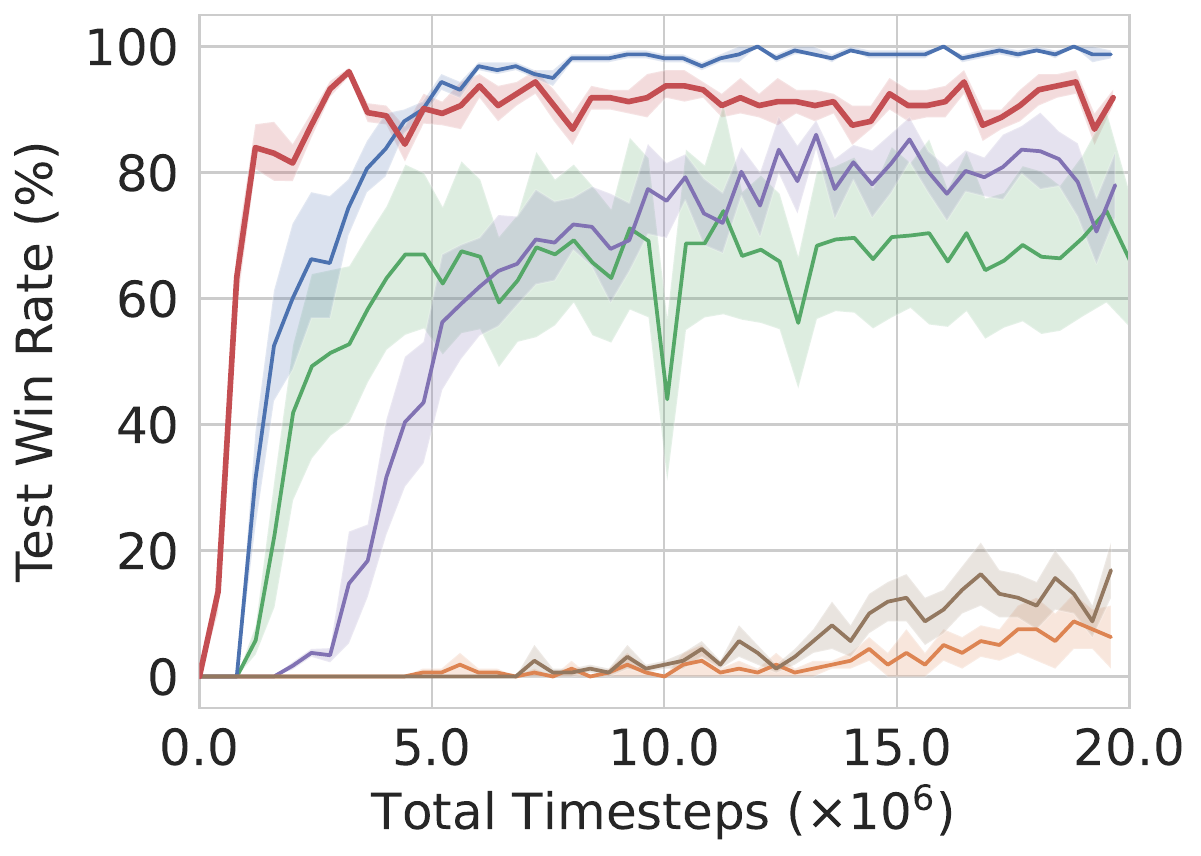}}\hfill
  \subfloat[6h\_vs\_8z (Super Hard)]{\includegraphics[width=0.333\textwidth]{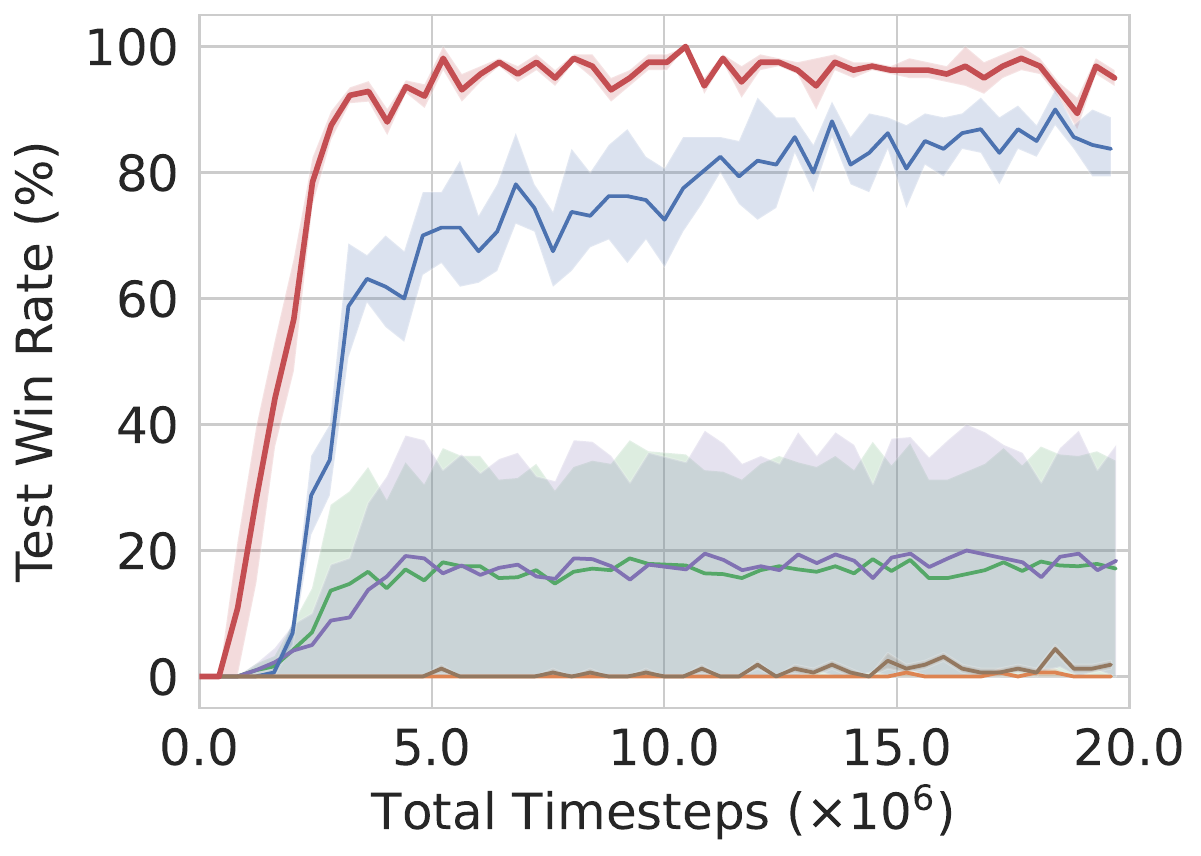}}\hfill
  \subfloat[3s5z\_vs\_3s6z (Super Hard)]{\includegraphics[width=0.333\textwidth]{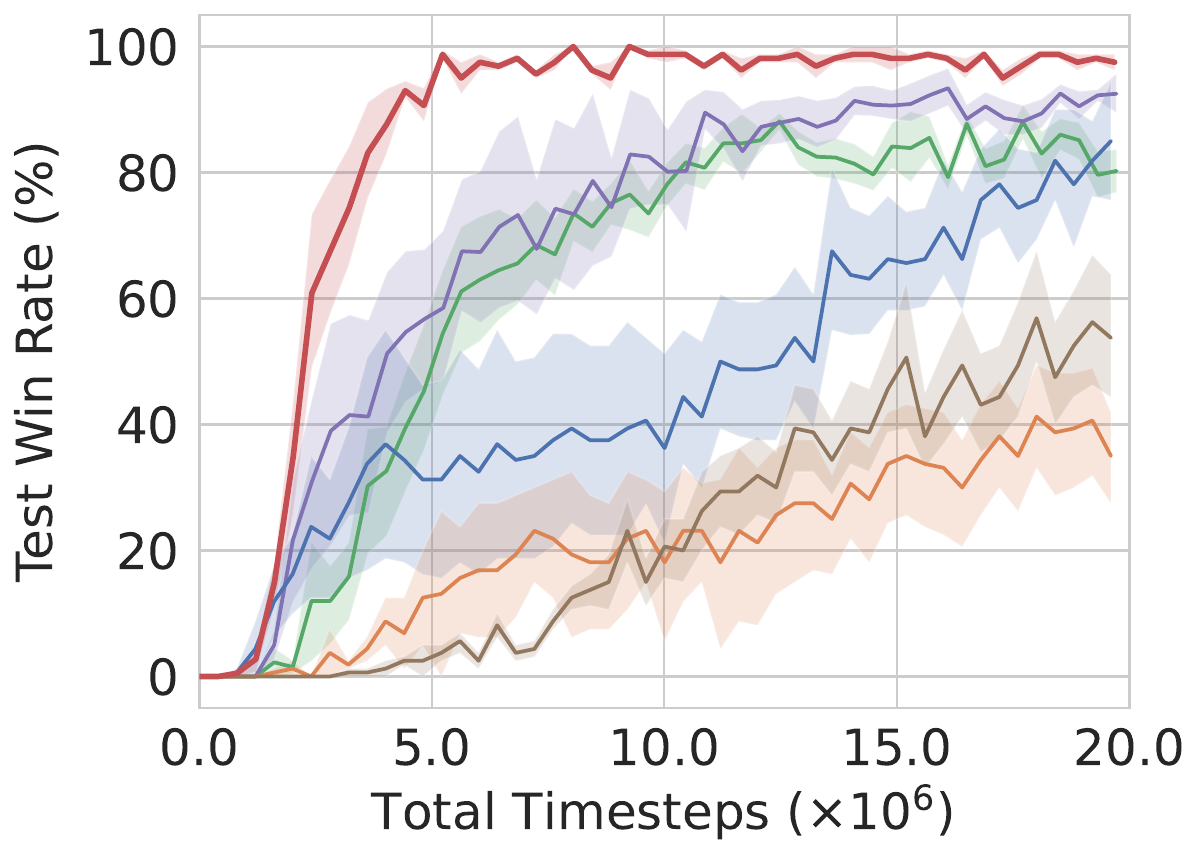}}
  \caption{Learning curves of our method and policy-based baselines in 6 single-task SMAC scenarios. We also provide the additional results of value-based QMIX and QPLEX using 20M training timesteps to ensure comparability.}
  \label{fig:sigle-task-20M}
  \vspace{-0.05cm}
\end{figure*}

\section{Experiments\label{sec:result}}

To demonstrate the effectiveness of the proposed interaction pattern disentangling (OPT) method, we conduct experiments on the StarCraft Multi-Agent Challenge~(SMAC)\footnote{We use SC2.4.10 version instead of the older SC2.4.6.2.69232. Performance is not comparable across versions.}~\cite{SMAC}, Google Research Football~(GRF)~\cite{GRF} and the Predator-Prey Game~(PPG)~\cite{mahajan2022generalization}. 
Our methods are compared with various state-of-the-art baseline methods in three settings: (1) single-task SMAC and GRF. (2) multi-task SMAC\footnote{We adopt the multi-task SMAC designed by Iqbal and Sha~\cite{REFIL}.}. (3) Zero-shot PPG.
We adopt the Python MARL framework (PyMARL)~\cite{SMAC} to run all value-based experiments and the On-policy framework~\cite{MAPPO} to run all policy-based experiments. The hyperparameters of the baselines are the same as those in their source codes. The implementation of our OPT method is also based on the above frameworks. In the SMAC and PPG experiments, we use QMIX~\cite{QMIX} as the backbone based on the PyMARL framework. In the GRF experiments, we use MAPPO~\cite{MAPPO} as the backbone based on the On-policy framework. The detailed hyperparameters are given as follows, where the common training parameters across different baselines are consistent to ensure comparability.

\begin{table*}[!t]
  \centering
  \caption{The area under curve and test win rate of our method and baselines in 6 single-task SMAC scenarios. $\pm$ corresponds to one standard deviation of the average evaluation over 5 trials. \textbf{Bold} indicates the best performance in each SMAC scenario. $\dagger$ indicates the results using 20M training timesteps as shown in Figure~\ref{fig:sigle-task-20M}.}
  \label{tab:sigle-task}\resizebox{\textwidth}{!}{%
  \begin{tabular}{@{}cccccccc@{}}
  \toprule
  \multicolumn{1}{l}{}                       & \textbf{Method} & \textbf{10m\_vs\_11m}  & \textbf{5m\_vs\_6m}    & \textbf{MMM2}          & \textbf{corridor}      & \textbf{6h\_vs\_8z}    & \textbf{3s5z\_vs\_3s6z} \\ \midrule
  \multirow{14}{*}{\textbf{Area Under Curve}} & \textbf{IQL}~\cite{SMAC}  & 0.09 $\pm$ 0.04  & 0.19 $\pm$ 0.03  & 0.00 $\pm$ 0.00  & 0.01 $\pm$ 0.01  & 0.00 $\pm$ 0.00  & 0.00 $\pm$ 0.00 \\ \specialrule{0em}{1pt}{1pt}
& \textbf{VDN}~\cite{VDN}  & 0.55 $\pm$ 0.02  & 0.44 $\pm$ 0.04  & 0.03 $\pm$ 0.03  & 0.05 $\pm$ 0.06  & 0.00 $\pm$ 0.00  & 0.05 $\pm$ 0.04 \\ \specialrule{0em}{1pt}{1pt}
& \textbf{QMIX}~\cite{QMIX}  & 0.61 $\pm$ 0.12  & 0.33 $\pm$ 0.10  & 0.17 $\pm$ 0.12  & 0.02 $\pm$ 0.02  & 0.12 $\pm$ 0.10  & 0.11 $\pm$ 0.19 \\ \specialrule{0em}{1pt}{1pt}
& \textbf{QTRAN}~\cite{QTRAN}  & 0.21 $\pm$ 0.08  & 0.27 $\pm$ 0.08  & 0.01 $\pm$ 0.01  & 0.00 $\pm$ 0.00  & 0.00 $\pm$ 0.00  & 0.00 $\pm$ 0.00 \\ \specialrule{0em}{1pt}{1pt}
& \textbf{QPLEX}~\cite{QPLEX}  & 0.50 $\pm$ 0.06  & 0.48 $\pm$ 0.09  & 0.05 $\pm$ 0.02  & 0.00 $\pm$ 0.00  & 0.05 $\pm$ 0.09  & 0.01 $\pm$ 0.02 \\ \specialrule{0em}{1pt}{1pt}
& \textbf{OWQMIX}~\cite{WQMIX}  & 0.47 $\pm$ 0.10  & 0.41 $\pm$ 0.08  & 0.28 $\pm$ 0.11  & 0.00 $\pm$ 0.00  & 0.29 $\pm$ 0.17  & 0.09 $\pm$ 0.08 \\ \specialrule{0em}{1pt}{1pt}
& \textbf{CWQMIX}~\cite{WQMIX}  & 0.49 $\pm$ 0.08  & 0.36 $\pm$ 0.09  & 0.24 $\pm$ 0.11  & 0.00 $\pm$ 0.00  & 0.23 $\pm$ 0.17  & 0.05 $\pm$ 0.04 \\ \specialrule{0em}{1pt}{1pt}\cmidrule(l){2-8}
& \textbf{Ours}  & \textbf{0.68 $\pm$ 0.04}  & \textbf{0.61 $\pm$ 0.05}  & \textbf{0.54 $\pm$ 0.08}  & \textbf{0.59 $\pm$ 0.05}  & \textbf{0.53 $\pm$ 0.12}  & \textbf{0.38 $\pm$ 0.13} \\ \specialrule{0em}{1pt}{1pt} \cmidrule(l){2-8}
& \textbf{QMIX$^\dagger$}~\cite{QMIX}  & 0.93 $\pm$ 0.01  & 0.67 $\pm$ 0.06  & 0.83 $\pm$ 0.02  & 0.59 $\pm$ 0.31  & 0.15 $\pm$ 0.30  & 0.62 $\pm$ 0.10 \\ \specialrule{0em}{1pt}{1pt}
& \textbf{QPLEX$^\dagger$}~\cite{QPLEX}  & 0.82 $\pm$ 0.03  & 0.75 $\pm$ 0.05  & 0.83 $\pm$ 0.02  & 0.60 $\pm$ 0.14  & 0.15 $\pm$ 0.30  & 0.69 $\pm$ 0.17 \\ \specialrule{0em}{1pt}{1pt}
& \textbf{MAPPO$^\dagger$}~\cite{MAPPO}  & 0.90 $\pm$ 0.04  & 0.77 $\pm$ 0.03  & 0.82 $\pm$ 0.04  & 0.88 $\pm$ 0.03  & 0.67 $\pm$ 0.12  & 0.46 $\pm$ 0.29 \\ \specialrule{0em}{1pt}{1pt}
& \textbf{HAPPO$^\dagger$}~\cite{HATRPO}  & 0.88 $\pm$ 0.06  & 0.02 $\pm$ 0.00  & 0.42 $\pm$ 0.24  & 0.02 $\pm$ 0.03  & 0.00 $\pm$ 0.00  & 0.20 $\pm$ 0.21 \\ \specialrule{0em}{1pt}{1pt}
& \textbf{HATRPO$^\dagger$}~\cite{HATRPO}  & 0.91 $\pm$ 0.01  & 0.70 $\pm$ 0.04  & 0.60 $\pm$ 0.14  & 0.05 $\pm$ 0.04  & 0.01 $\pm$ 0.00  & 0.23 $\pm$ 0.12 \\ \specialrule{0em}{1pt}{1pt}\cmidrule(l){2-8}
& \textbf{Ours$^\dagger$}  & \textbf{0.96 $\pm$ 0.00}  & \textbf{0.86 $\pm$ 0.03}  & \textbf{0.94 $\pm$ 0.01}  & \textbf{0.89 $\pm$ 0.05}  & \textbf{0.87 $\pm$ 0.02}  & \textbf{0.85 $\pm$ 0.04} \\ \specialrule{0em}{1pt}{1pt}\midrule 
\multirow{14}{*}{\textbf{Test Win Rate}} & \textbf{IQL}~\cite{SMAC}  & 0.26 $\pm$ 0.16  & 0.47 $\pm$ 0.09  & 0.00 $\pm$ 0.00  & 0.08 $\pm$ 0.15  & 0.00 $\pm$ 0.00  & 0.00 $\pm$ 0.00 \\ \specialrule{0em}{1pt}{1pt}
& \textbf{VDN}~\cite{VDN}  & 0.94 $\pm$ 0.04  & 0.73 $\pm$ 0.05  & 0.20 $\pm$ 0.27  & 0.25 $\pm$ 0.21  & 0.00 $\pm$ 0.00  & 0.27 $\pm$ 0.19 \\ \specialrule{0em}{1pt}{1pt}
& \textbf{QMIX}~\cite{QMIX}  & 0.88 $\pm$ 0.13  & 0.57 $\pm$ 0.19  & 0.54 $\pm$ 0.30  & 0.20 $\pm$ 0.18  & 0.37 $\pm$ 0.31  & 0.25 $\pm$ 0.34 \\ \specialrule{0em}{1pt}{1pt}
& \textbf{QTRAN}~\cite{QTRAN}  & 0.60 $\pm$ 0.23  & 0.50 $\pm$ 0.20  & 0.02 $\pm$ 0.02  & 0.00 $\pm$ 0.00  & 0.00 $\pm$ 0.00  & 0.00 $\pm$ 0.00 \\ \specialrule{0em}{1pt}{1pt}
& \textbf{QPLEX}~\cite{QPLEX}  & 0.79 $\pm$ 0.08  & 0.70 $\pm$ 0.09  & 0.44 $\pm$ 0.14  & 0.01 $\pm$ 0.01  & 0.13 $\pm$ 0.27  & 0.08 $\pm$ 0.14 \\ \specialrule{0em}{1pt}{1pt}
& \textbf{OWQMIX}~\cite{WQMIX}  & 0.75 $\pm$ 0.14  & 0.54 $\pm$ 0.18  & 0.62 $\pm$ 0.27  & 0.00 $\pm$ 0.00  & 0.54 $\pm$ 0.27  & 0.34 $\pm$ 0.23 \\ \specialrule{0em}{1pt}{1pt}
& \textbf{CWQMIX}~\cite{WQMIX}  & 0.77 $\pm$ 0.12  & 0.56 $\pm$ 0.11  & 0.69 $\pm$ 0.19  & 0.00 $\pm$ 0.00  & 0.35 $\pm$ 0.29  & 0.16 $\pm$ 0.13 \\ \specialrule{0em}{1pt}{1pt}\cmidrule(l){2-8} 
& \textbf{Ours}  & \textbf{0.97 $\pm$ 0.02}  & \textbf{0.86 $\pm$ 0.07}  & \textbf{0.96 $\pm$ 0.03}  & \textbf{0.88 $\pm$ 0.08}  & \textbf{0.94 $\pm$ 0.03}  & \textbf{0.84 $\pm$ 0.17} \\ \specialrule{0em}{1pt}{1pt}\cmidrule(l){2-8} 
& \textbf{QMIX$^\dagger$}~\cite{QMIX}  & 0.96 $\pm$ 0.02  & 0.70 $\pm$ 0.12  & 0.94 $\pm$ 0.02  & 0.66 $\pm$ 0.40  & 0.19 $\pm$ 0.39  & 0.80 $\pm$ 0.11 \\ \specialrule{0em}{1pt}{1pt}
& \textbf{QPLEX$^\dagger$}~\cite{QPLEX}  & 0.88 $\pm$ 0.05  & 0.82 $\pm$ 0.09  & 0.98 $\pm$ 0.03  & 0.79 $\pm$ 0.14  & 0.15 $\pm$ 0.30  & 0.94 $\pm$ 0.05 \\ \specialrule{0em}{1pt}{1pt}
& \textbf{MAPPO$^\dagger$}~\cite{MAPPO}  & 0.99 $\pm$ 0.02  & 0.86 $\pm$ 0.06  & 0.96 $\pm$ 0.02  & \textbf{0.99 $\pm$ 0.02}  & 0.89 $\pm$ 0.06  & 0.74 $\pm$ 0.28 \\ \specialrule{0em}{1pt}{1pt}
& \textbf{HAPPO$^\dagger$}~\cite{HATRPO}  & 0.96 $\pm$ 0.01  & 0.06 $\pm$ 0.02  & 0.65 $\pm$ 0.24  & 0.07 $\pm$ 0.09  & 0.01 $\pm$ 0.01  & 0.39 $\pm$ 0.29 \\ \specialrule{0em}{1pt}{1pt}
& \textbf{HATRPO$^\dagger$}~\cite{HATRPO}  & 0.99 $\pm$ 0.02  & 0.82 $\pm$ 0.02  & 0.91 $\pm$ 0.10  & 0.13 $\pm$ 0.11  & 0.03 $\pm$ 0.02  & 0.54 $\pm$ 0.27 \\ \specialrule{0em}{1pt}{1pt}\cmidrule(l){2-8} 
& \textbf{Ours$^\dagger$}  & \textbf{0.99 $\pm$ 0.01}  & \textbf{0.89 $\pm$ 0.03}  & \textbf{0.98 $\pm$ 0.02}  & 0.92 $\pm$ 0.05  & \textbf{0.94 $\pm$ 0.03}  & \textbf{0.96 $\pm$ 0.04} \\ \specialrule{0em}{1pt}{1pt}\bottomrule 
  \end{tabular}}
  \vspace{-0.1cm}
\end{table*}

For the OPT module in all experiments, we set the layer number $K$ to 2, the interaction prototype number $N$ to 4, and the interaction prototype dimension $d_x$ to 32. The CD loss coefficient $\alpha$ is set to 0.5, and the CMI loss coefficient $\beta$ is set to 0.1. In the value-based experiments, batches of 32 episodes are sampled from the replay buffer with the size of 5K every training iteration. The target update interval is set to 200, and the discount factor is set to 0.99. We use the RMSprop Optimizer with a learning rate of $5 \times 10^{-4}$, a smoothing constant of 0.99, and no momentum or weight decay. For exploration, $\epsilon$-greedy is used with $\epsilon$ annealed linearly from 1.0 to 0.05 over 50K training steps and kept constant for the rest of the training. We modify several hyperparameters of the value-based experiments in some difficult scenarios and detail them below. In the policy-based experiments, MAPPO has performed large-scale hyperparameter searching for different scenarios~\cite{MAPPO}. 
Thus, we directly adopt the same hyperparameters as in MAPPO to conduct policy-based experiments. 
In the following, we will first introduce the environments and the baselines, and then the results of different methods and ablations.

\subsection{Single-task Environments}

We conduct experiments on 6 single-task SMAC scenarios~\cite{SMAC} and 3 single-task GRF scenarios~\cite{GRF}. The SMAC scenarios are classified into \textbf{Easy}~(\emph{10m\_vs\_11m}), \textbf{Hard}~(\emph{5m\_vs\_6m}), and \textbf{Super Hard}~(\emph{MMM2}, \emph{corridor}, \emph{6h\_vs\_8z}, \emph{3s5z\_vs\_3s6z}). Two of these scenarios are homogeneous~(\emph{10m\_vs\_11m}, \emph{5m\_vs\_6m}), where the army is composed of only a single unit type, while the others are heterogeneous. The GRF scenarios consists of \emph{academy\_3\_vs\_1\_with\_keeper}, \emph{academy\_counterattack\_easy}, and \emph{academy\_counterattack\_hard}. The goal of the agents is to score a goal against the opponent team. In the \emph{academy\_counterattack\_easy} and \emph{academy\_counterattack\_hard} scenarios, we follow the dense-reward setting~\cite{MAPPO} to reward agents for moving the ball closer to the opponent goal. In the \emph{academy\_3\_vs\_1\_with\_keeper} scenarios, to increase the learning difficulty, we adopt the sparse-reward setting where only scoring leads to the rewards.

\begin{figure*}[!t]
  \centering
  \subfloat{\quad\quad\includegraphics[scale=0.8]{fig/sc2_20M_legend.pdf}}\\    
    \addtocounter{subfigure}{-1}
    \vspace{-0.1cm}
  \subfloat[academy\_3\_vs\_1\_with\_keeper]{\includegraphics[width=0.333\textwidth]{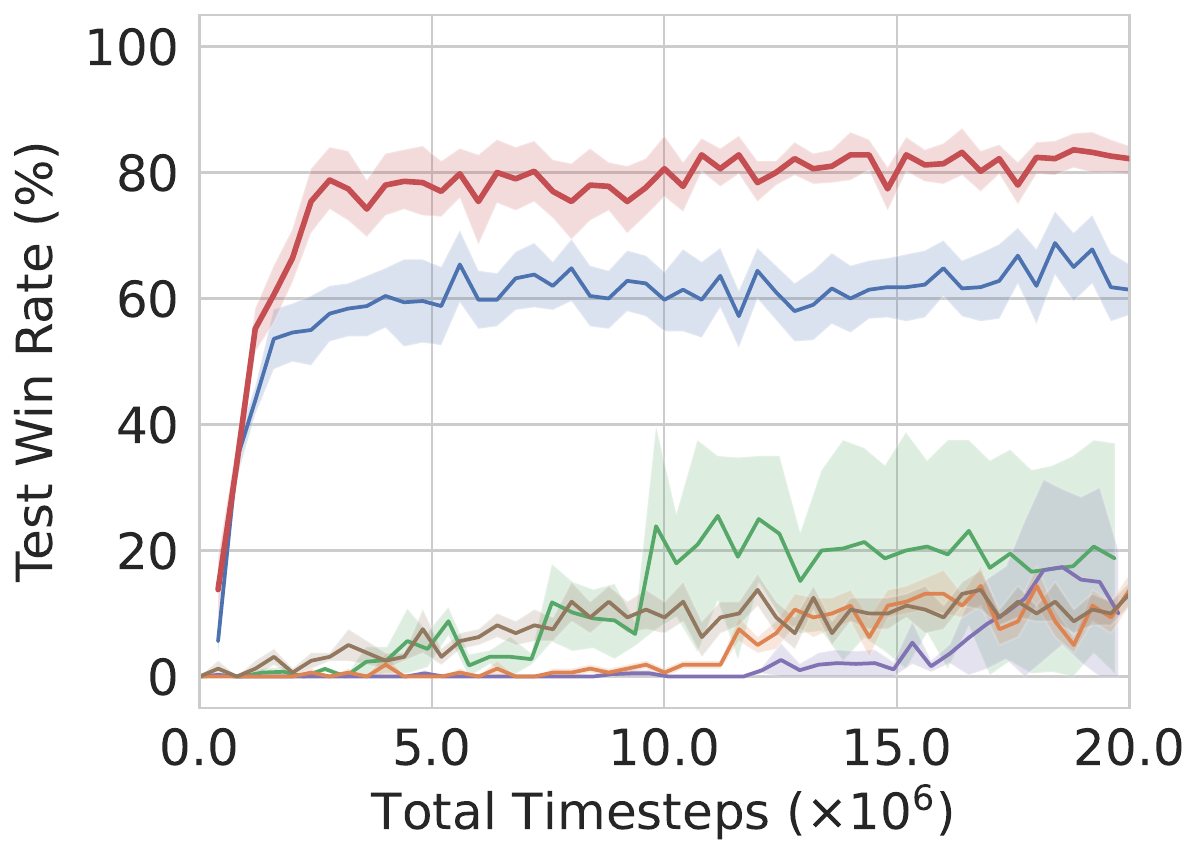}}\hfill
  \subfloat[academy\_counterattack\_easy]{\includegraphics[width=0.333\textwidth]{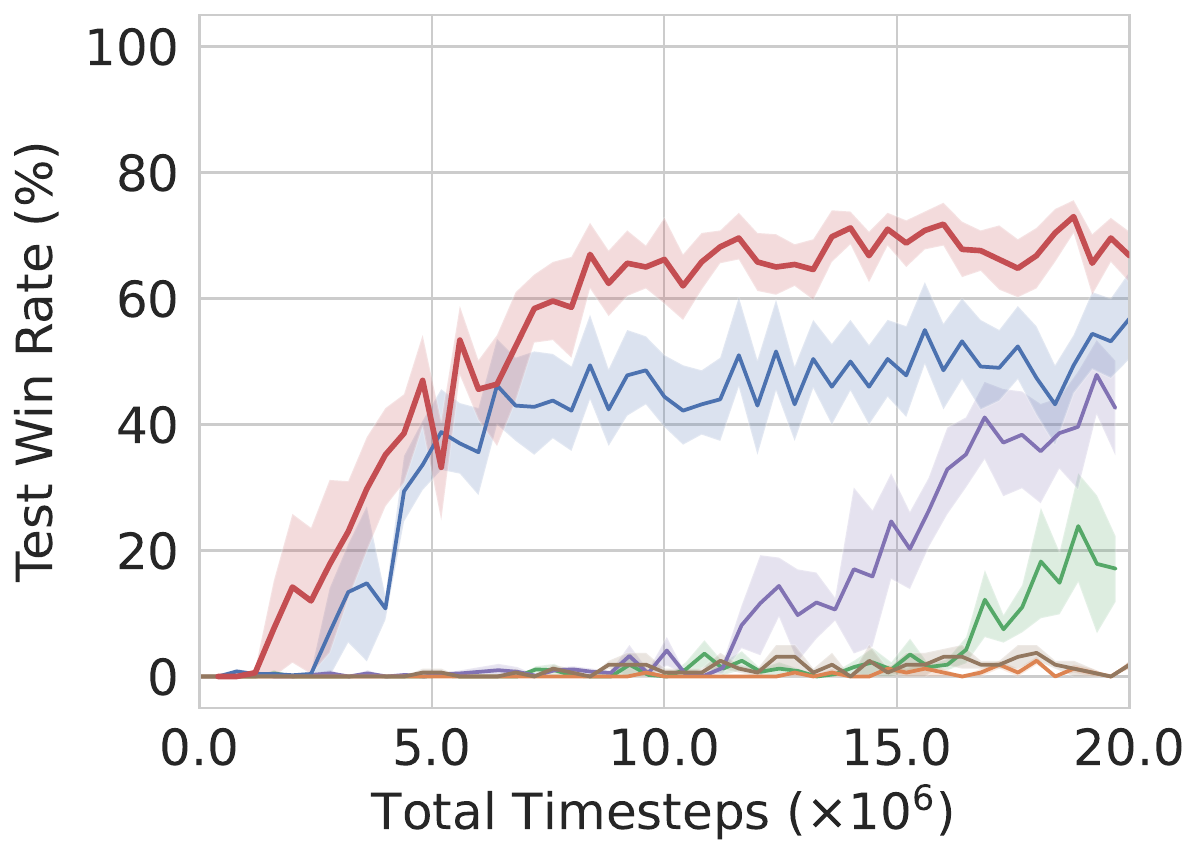}}\hfill
  \subfloat[academy\_counterattack\_hard]{\includegraphics[width=0.333\textwidth]{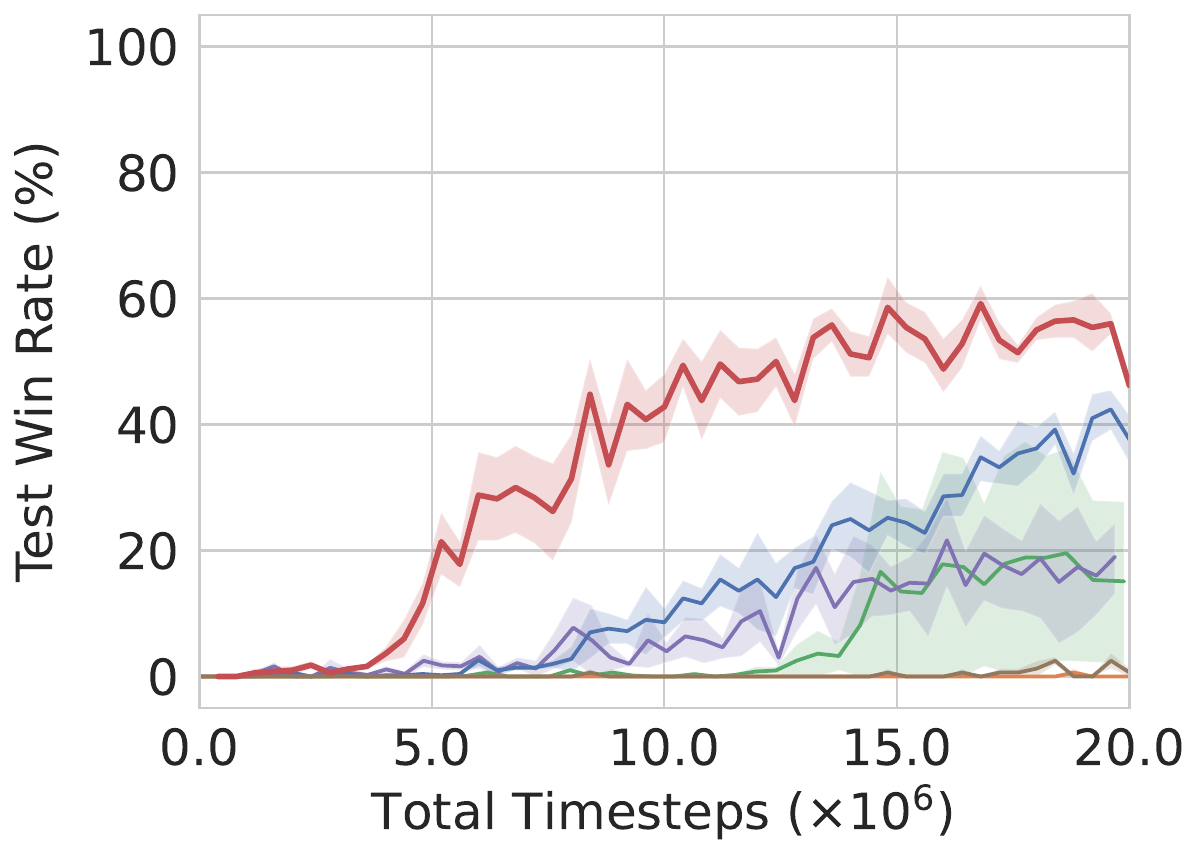}}
  \caption{Learning curves of our method and baselines in 3 single-task GRF scenarios.}
  \label{fig:gfootball}
  \vspace{-0.2cm}
\end{figure*}

\begin{table}[!t]
  \centering
  \caption{The area under curve and test win rate of our method and baselines in 3 single-task GRF scenarios.}
  \label{tab:gfootball}
  \resizebox{0.48\textwidth}{!}{%
  \begin{tabular}{@{}cccc@{}}
  \toprule
  \multirow{1}{*}{\textbf{Method}} & \textbf{3 vs 1}  & \textbf{CA (easy)}    & \textbf{CA (hard)}    \\ \midrule
  & \multicolumn{3}{c}{\textbf{Area Under Curve}} \\ \cmidrule(l){2-4} 
\textbf{QMIX}~\cite{QMIX}  & 0.12 $\pm$ 0.20  & 0.04 $\pm$ 0.04  & 0.05 $\pm$ 0.09 \\ \specialrule{0em}{1pt}{1pt}
\textbf{QPLEX}~\cite{QPLEX}   & 0.03 $\pm$ 0.05  & 0.13 $\pm$ 0.08  & 0.08 $\pm$ 0.09 \\ \specialrule{0em}{1pt}{1pt}
\textbf{MAPPO}~\cite{MAPPO}   & 0.59 $\pm$ 0.13  & 0.37 $\pm$ 0.16  & 0.13 $\pm$ 0.05 \\ \specialrule{0em}{1pt}{1pt}
\textbf{HAPPO}~\cite{HATRPO}   & 0.04 $\pm$ 0.03  & 0.00 $\pm$ 0.00  & 0.00 $\pm$ 0.00 \\ \specialrule{0em}{1pt}{1pt}
\textbf{HATRPO}~\cite{HATRPO}   & 0.08 $\pm$ 0.05  & 0.01 $\pm$ 0.02  & 0.00 $\pm$ 0.00 \\ \specialrule{0em}{1pt}{1pt}\midrule
\textbf{Ours}  & \textbf{0.75 $\pm$ 0.11}  & \textbf{0.52 $\pm$ 0.13}  & \textbf{0.34 $\pm$ 0.08} \\ \specialrule{0em}{1pt}{1pt}\midrule 
& \multicolumn{3}{c}{\textbf{Test Win Rate}} \\ \cmidrule(l){2-4} 
\textbf{QMIX}~\cite{QMIX}   & 0.20 $\pm$ 0.36  & 0.26 $\pm$ 0.32  & 0.14 $\pm$ 0.27 \\ \specialrule{0em}{1pt}{1pt}
\textbf{QPLEX}~\cite{QPLEX}   & 0.14 $\pm$ 0.28  & 0.54 $\pm$ 0.20  & 0.25 $\pm$ 0.25 \\ \specialrule{0em}{1pt}{1pt}
\textbf{MAPPO}~\cite{MAPPO}   & 0.61 $\pm$ 0.11  & 0.57 $\pm$ 0.22  & 0.38 $\pm$ 0.10 \\ \specialrule{0em}{1pt}{1pt}
\textbf{HAPPO}~\cite{HATRPO}   & 0.14 $\pm$ 0.08  & 0.02 $\pm$ 0.02  & 0.00 $\pm$ 0.00 \\ \specialrule{0em}{1pt}{1pt}
\textbf{HATRPO}~\cite{HATRPO}   & 0.13 $\pm$ 0.06  & 0.02 $\pm$ 0.02  & 0.01 $\pm$ 0.01 \\ \specialrule{0em}{1pt}{1pt}\midrule
\textbf{Ours}  & \textbf{0.82 $\pm$ 0.07}  & \textbf{0.67 $\pm$ 0.12}  & \textbf{0.46 $\pm$ 0.05} \\ \specialrule{0em}{1pt}{1pt}\bottomrule 
  \end{tabular}}
  \vspace{-0.2cm}
\end{table}

We compare the proposed OPT with various MARL baselines: (1) Value-based methods: IQL~\cite{SMAC}, VDN~\cite{VDN}, QMIX~\cite{QMIX}, QTRAN~\cite{QTRAN}, QPLEX~\cite{QPLEX}, OWQMIX and CWQMIX~\cite{WQMIX}. (2) Policy-based methods: MAPPO~\cite{MAPPO}, HAPPO and HATRPO~\cite{HATRPO}. For the value-based methods, the training timestep is set to 2M or 4M in different scenarios. For the policy-based methods, the training timestep is set to 20M due to the low sample efficiency. We also provide the additional results of value-based QMIX and QPLEX using 20M training timesteps for comparison. For the super hard SMAC scenarios that require more exploration, the test win rates of value-based methods often remain $0\%$ with the default hyperparameters. In this way, we extend the epsilon anneal time to 500K, and two of them~(\emph{corridor}, \emph{6h\_vs\_8z}) optimized with the Adam Optimizer for all the compared value-based methods.

The experimental results in SMAC compared with the state-of-the-art value-based methods are shown in Figure~\ref{fig:sigle-task} and Table~\ref{tab:sigle-task}. In both easy and hard homogeneous scenarios~(\emph{10m\_vs\_11m}, \emph{5m\_vs\_6m}), our proposed OPT successfully improves the learning efficiency and the final performance. In the easy homogeneous scenario~(\emph{10m\_vs\_11m}), the size of the armed forces between the agents and enemies is similar. Thus, several baselines, including VDN and QMIX, can also achieve the promising results with additive or monotonic value factorization, while the exploratory benefit brought by interaction pattern disentangling is not obvious. However, in the more difficult homogeneous scenario (\emph{5m\_vs\_6m}), our method consistently outperforms baselines by a large margin during training. The results suggest that the disentangling can be utilized to explore the diverse cooperation mechanisms, which helps the agents to construct a more useful policy and achieve non-trivial performance.

\begin{figure*}[!t]
  \centering
  \subfloat{\quad\quad\includegraphics[scale=0.8]{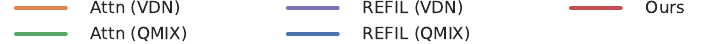}}\\    
    \addtocounter{subfigure}{-1}
  \vspace{-0.1cm}
  \subfloat[3-8csz (66 Tasks, Easy)]{\includegraphics[width=0.333\textwidth]{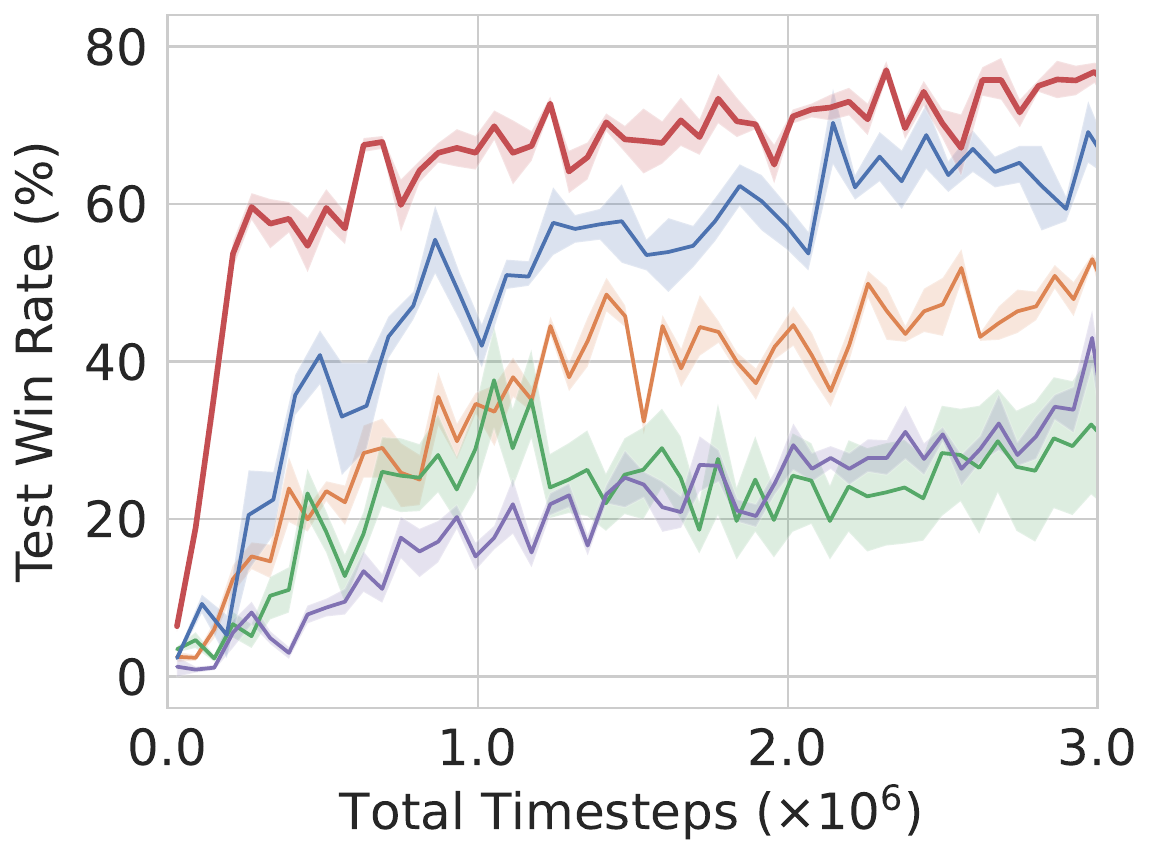}}\hfill
  \subfloat[3-8MMM (66 Tasks, Hard)]{\includegraphics[width=0.333\textwidth]{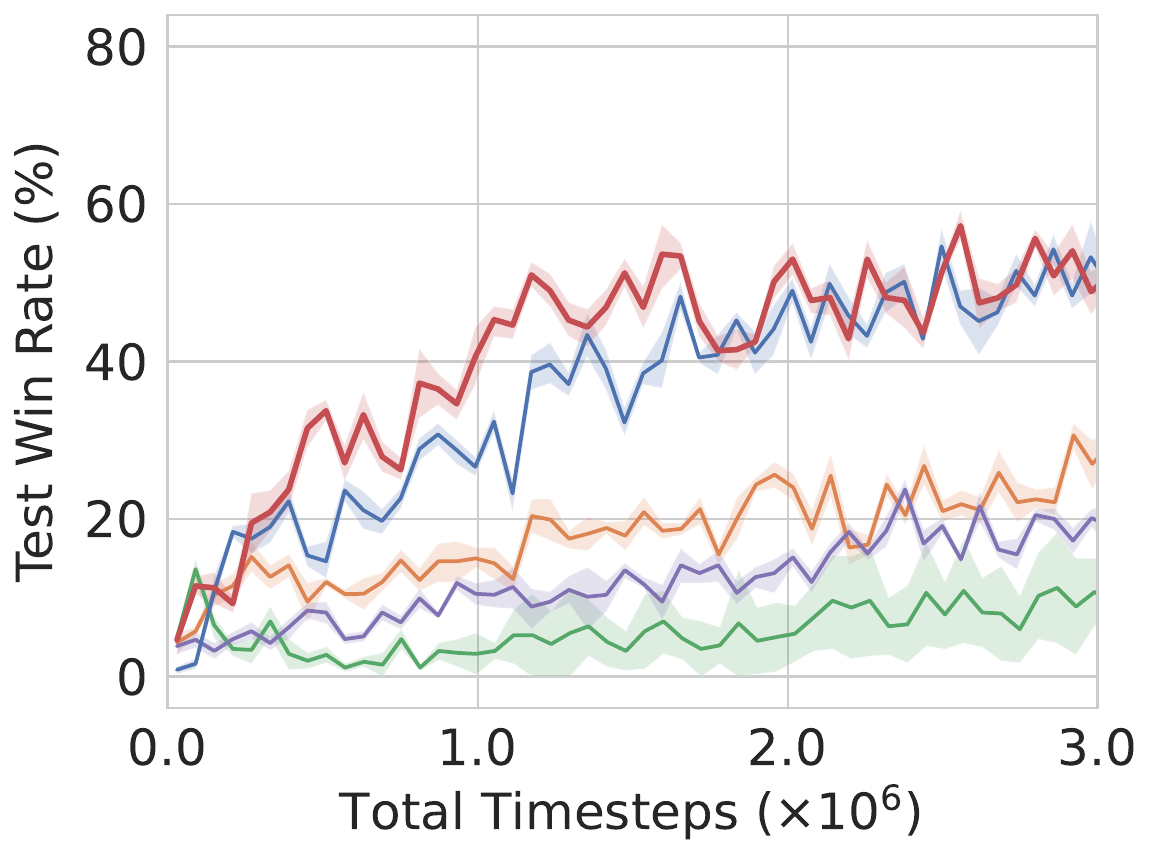}}\hfill
  \subfloat[3-8sz (39 Tasks, Hard)]{\includegraphics[width=0.333\textwidth]{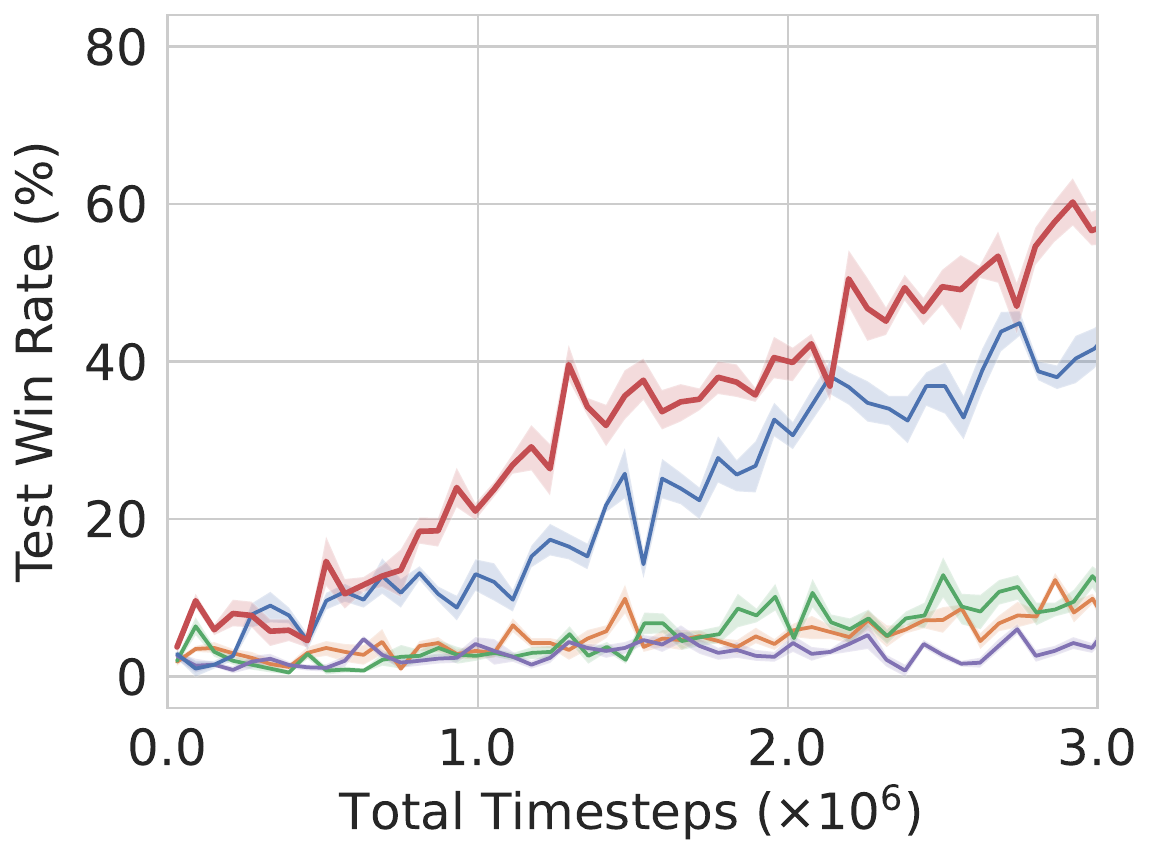}}
  
  \subfloat[5-11csz (120 Tasks, Easy)]{\includegraphics[width=0.333\textwidth]{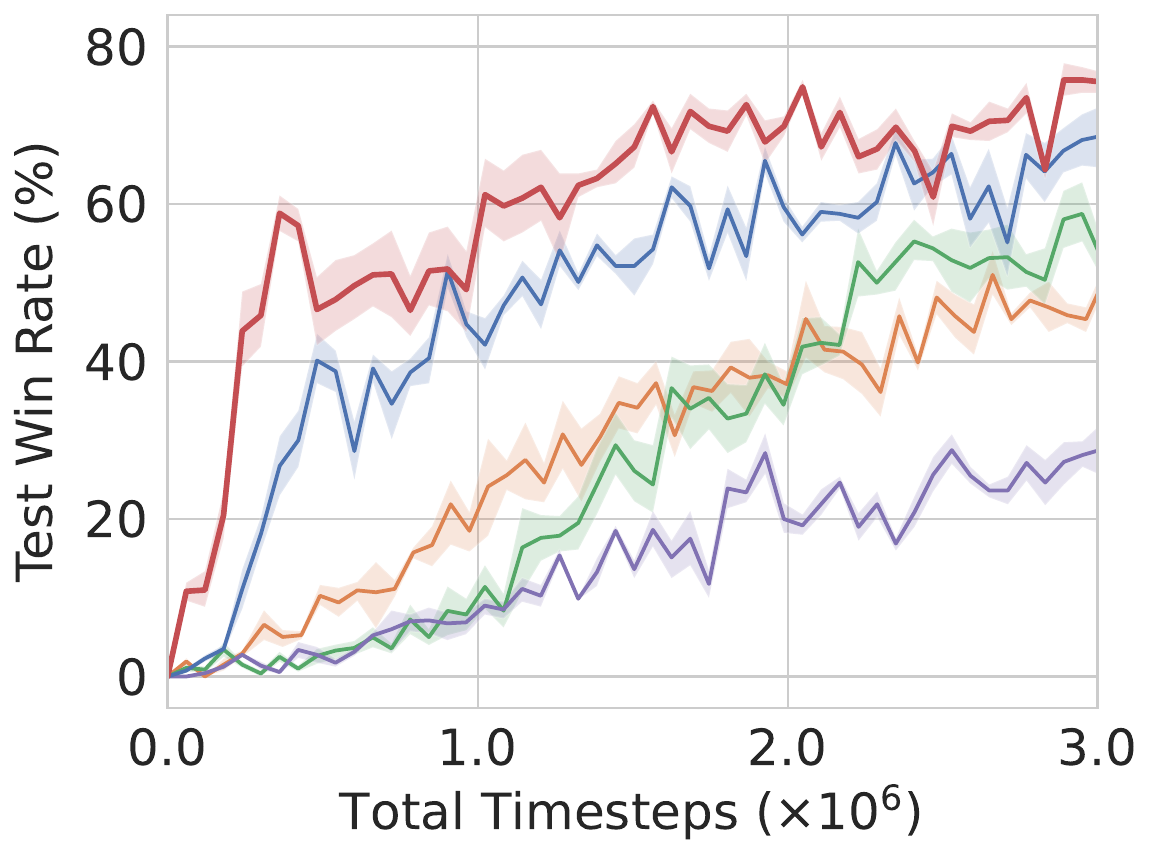}}\hfill
  \subfloat[5-11MMM (120 Tasks, Super Hard)]{\includegraphics[width=0.333\textwidth]{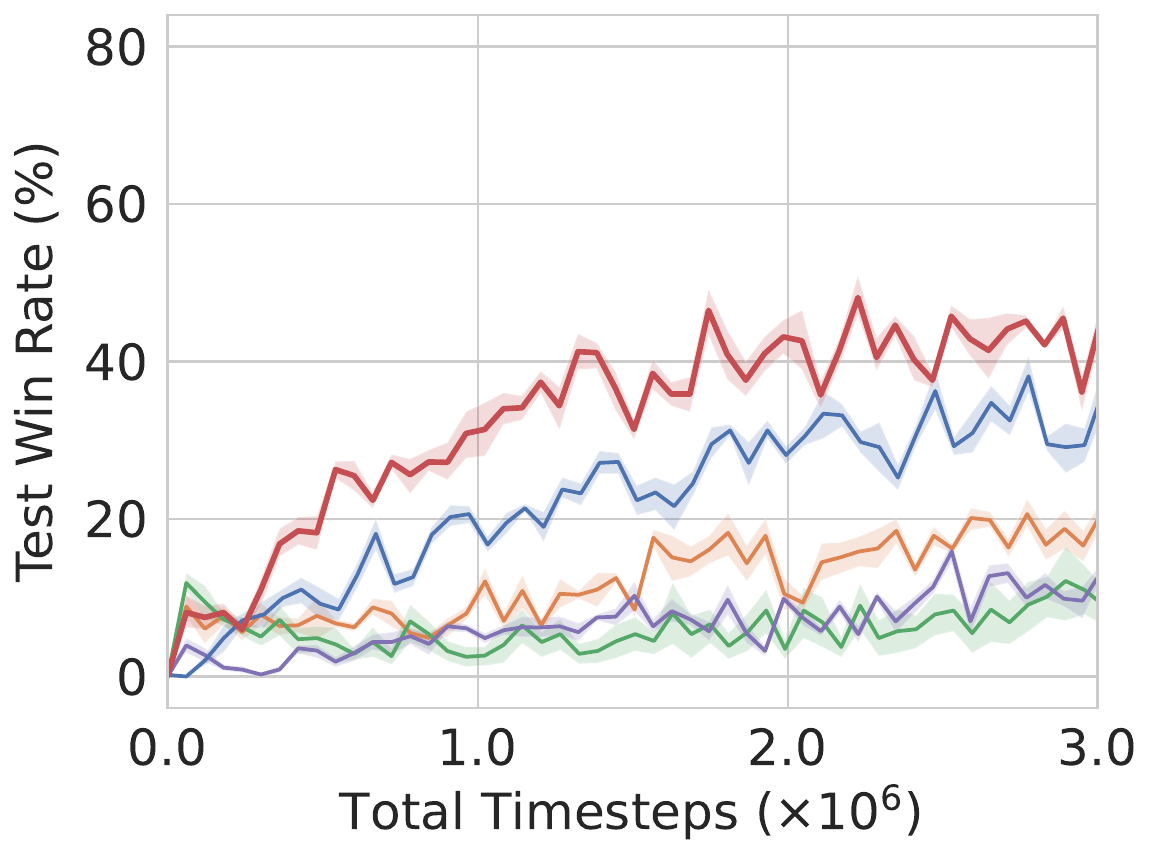}}\hfill
  \subfloat[5-11sz (63 Tasks, Super Hard)]{\includegraphics[width=0.333\textwidth]{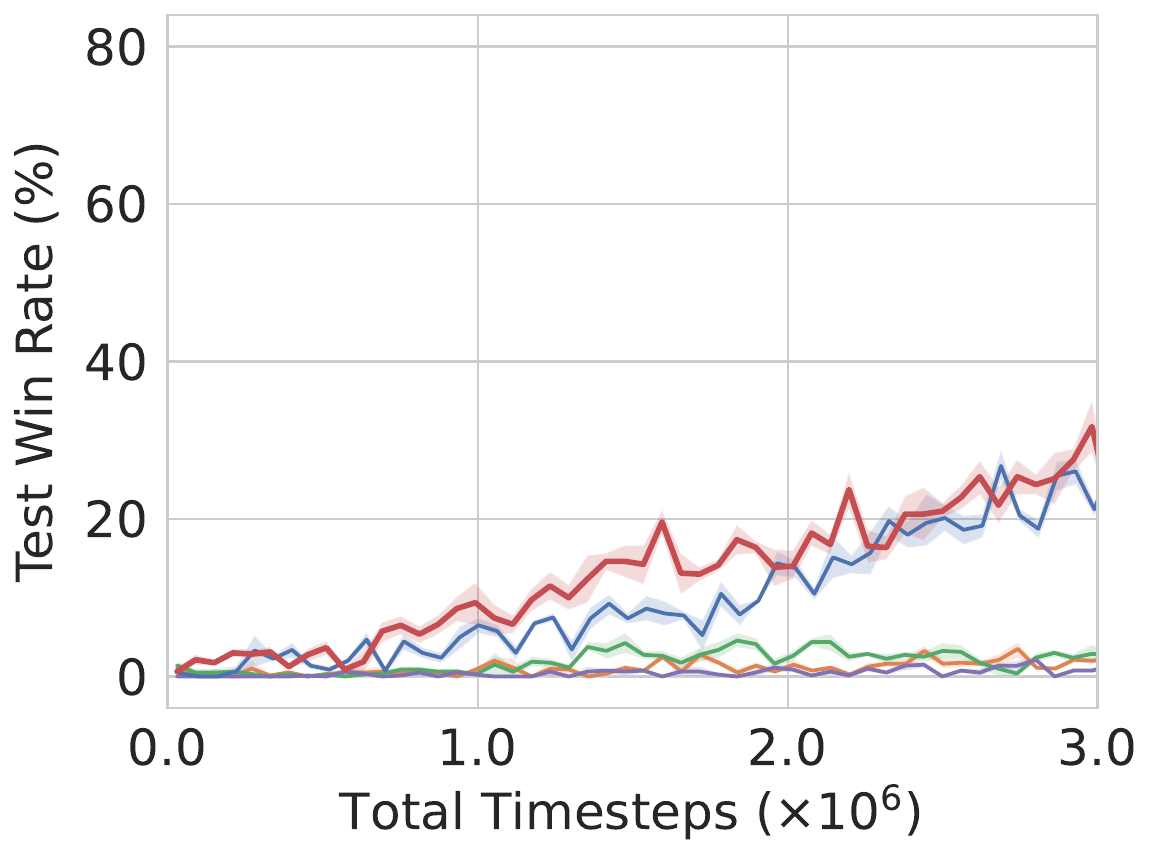}}
  \caption{Learning curves of our method and baselines in 6 multi-task SMAC scenarios.}
  \label{fig:multi-task}
\end{figure*}

\begin{table*}[!t]
  \centering
  \caption{The area under curve and test win rate of our method and baselines in 6 multi-task SMAC scenarios.}
  \label{tab:multi-task}
  \resizebox{\textwidth}{!}{%
  \begin{tabular}{@{}cccccccc@{}}
  \toprule
  \multicolumn{1}{l}{}                       & \textbf{Method} & \textbf{3-8csz}  & \textbf{3-8MMM}    & \textbf{3-8sz}          & \textbf{5-11csz}      & \textbf{5-11MMM}    & \textbf{5-11sz} \\ \midrule
  \multirow{5}{*}{\textbf{Area Under Curve}} & \textbf{Attn (VDN)}~\cite{VDN}  & 0.36 $\pm$ 0.03  & 0.18 $\pm$ 0.02  & 0.05 $\pm$ 0.01  & 0.29 $\pm$ 0.05  & 0.12 $\pm$ 0.01  & 0.01 $\pm$ 0.00 \\ \specialrule{0em}{1pt}{1pt}
& \textbf{Attn (QMIX)}~\cite{QMIX}  & 0.22 $\pm$ 0.13  & 0.06 $\pm$ 0.08  & 0.05 $\pm$ 0.02  & 0.27 $\pm$ 0.06  & 0.06 $\pm$ 0.05  & 0.02 $\pm$ 0.00 \\ \specialrule{0em}{1pt}{1pt}
& \textbf{REFIL (VDN)}~\cite{REFIL}  & 0.21 $\pm$ 0.03  & 0.12 $\pm$ 0.01  & 0.03 $\pm$ 0.01  & 0.14 $\pm$ 0.01  & 0.07 $\pm$ 0.01  & 0.00 $\pm$ 0.00 \\ \specialrule{0em}{1pt}{1pt}
& \textbf{REFIL (QMIX)}~\cite{REFIL}  & 0.51 $\pm$ 0.05  & 0.35 $\pm$ 0.01  & 0.23 $\pm$ 0.02  & 0.48 $\pm$ 0.02  & 0.22 $\pm$ 0.01  & 0.10 $\pm$ 0.01 \\ \specialrule{0em}{1pt}{1pt}\cmidrule(l){2-8}
& \textbf{Ours}  & \textbf{0.65 $\pm$ 0.02}  & \textbf{0.40 $\pm$ 0.03}  & \textbf{0.32 $\pm$ 0.03}  & \textbf{0.60 $\pm$ 0.03}  & \textbf{0.33 $\pm$ 0.02}  & \textbf{0.13 $\pm$ 0.02} \\ \specialrule{0em}{1pt}{1pt}\midrule 
\multirow{5}{*}{\textbf{Test Win Rate}} & \textbf{Attn (VDN)}~\cite{VDN}  & 0.52 $\pm$ 0.10  & 0.24 $\pm$ 0.04  & 0.06 $\pm$ 0.02  & 0.42 $\pm$ 0.10  & 0.15 $\pm$ 0.02  & 0.01 $\pm$ 0.01 \\ \specialrule{0em}{1pt}{1pt}
& \textbf{Attn (QMIX)}~\cite{QMIX}  & 0.28 $\pm$ 0.27  & 0.08 $\pm$ 0.10  & 0.11 $\pm$ 0.07  & 0.56 $\pm$ 0.12  & 0.11 $\pm$ 0.09  & 0.01 $\pm$ 0.01 \\ \specialrule{0em}{1pt}{1pt}
& \textbf{REFIL (VDN)}~\cite{REFIL}  & 0.39 $\pm$ 0.06  & 0.20 $\pm$ 0.04  & 0.06 $\pm$ 0.03  & 0.29 $\pm$ 0.08  & 0.12 $\pm$ 0.05  & 0.00 $\pm$ 0.00 \\ \specialrule{0em}{1pt}{1pt}
& \textbf{REFIL (QMIX)}~\cite{REFIL}  & 0.66 $\pm$ 0.12  & 0.53 $\pm$ 0.07  & 0.44 $\pm$ 0.07  & 0.65 $\pm$ 0.08  & 0.27 $\pm$ 0.01  & 0.25 $\pm$ 0.05 \\ \specialrule{0em}{1pt}{1pt}\cmidrule(l){2-8} 
& \textbf{Ours}  & \textbf{0.79 $\pm$ 0.06}  & \textbf{0.53 $\pm$ 0.03}  & \textbf{0.57 $\pm$ 0.06}  & \textbf{0.75 $\pm$ 0.07}  & \textbf{0.47 $\pm$ 0.10}  & \textbf{0.27 $\pm$ 0.03} \\ \specialrule{0em}{1pt}{1pt}\bottomrule 
  \end{tabular}}
  \vspace{-0.2cm}
\end{table*}

To further test the potentiality of our proposed method, we compare our method in the super hard SMAC heterogeneous scenarios~(\emph{MMM2}, \emph{corridor}, \emph{6h\_vs\_8z}, \emph{3s5z\_vs\_3s6z}). In these challenging scenarios, extracting interaction patterns becomes complex due to the different unit types. Furthermore, there is a great disparity in strength between the two teams, and it is impossible to beat the enemies in a reckless way. The agents have to learn proper tactics to win the battle, while the sophisticated tactics are laborious to explore. In the \emph{MMM2} and \emph{6h\_vs\_8z} scenarios, OWQMIX and CWQMIX use a weighted projection that allows more emphasis to be placed on better joint actions, giving their improved ability in cooperative exploration. Even so, they inevitably fall into suboptimal policies with low test win rates. In contrast, the proposed OPT provides an impressive improvement in the performance over the baselines, showing its robustness to an increased rate and quality of exploration. Especially in the \emph{corridor} and \emph{3s5z\_vs\_3s6z} scenarios, almost all compared baselines cannot learn any effective policy and perform poorly. Our method still maintains a high learning efficiency, and in practice, leads to the superior performance compared to the baselines.

The experimental results in SMAC compared with the state-of-the-art policy-based methods are shown in Figure~\ref{fig:sigle-task-20M} and Table~\ref{tab:sigle-task}. In the easy and hard homogeneous scenarios, our proposed OPT can achieve gratifying performances on par with those obtained by the policy-based methods. Furthermore, OPT still provides competent efficiency and performance guarantees in all super hard scenarios except the \emph{corridor} scenario where the final performance of OPT is only inferior to MAPPO. Besides, the additional results of QMIX and QPLEX show that more training timesteps enable value-based methods to converge to a higher performance, while OPT  still maintains a better performance with a lower training cost. 

Figure~\ref{fig:gfootball} and Table~\ref{tab:gfootball} report the experimental results in GRF. In all scenarios, agents must learn diverse cooperative tactics to pass the ball and score the winning goal. Our proposed OPT successfully provides competent performance guarantees compared to the baselines. In the \emph{academy\_3\_vs\_1\_with\_keeper} scenario where agents are only rewarded for scoring, it is challenging for agents to find an effective policy due to the extreme sparsity of rewards. Thus, the compared baselines, except for MAPPO,  all perform poorly. Despite the promising results achieved by MAPPO, MAPPO is easy to fall into local optimum. In contrast, OPT can benefit from interaction pattern disentangling to explore a more generalized cooperative policy. In the \emph{academy\_counterattack\_easy} and \emph{academy\_counterattack\_hard} scenarios, the difficulty of exploration increases as the number of entities increases. However, OPT still shows its robustness to cooperation promoting and offers impressive performance. Overall, these extensive results exhibit the advantage of boosting interaction pattern disentangling in single-task MARL, yielding significantly superior performance compared to the state-of-the-art counterparts.

\begin{figure*}[!t]
  \centering
  \subfloat{\quad\quad\includegraphics[scale=0.8]{fig/mt_legend.pdf}}\\    
    \addtocounter{subfigure}{-1}
  \vspace{-0.1cm}
  \subfloat[Unseen Capability]{\includegraphics[width=0.333\textwidth]{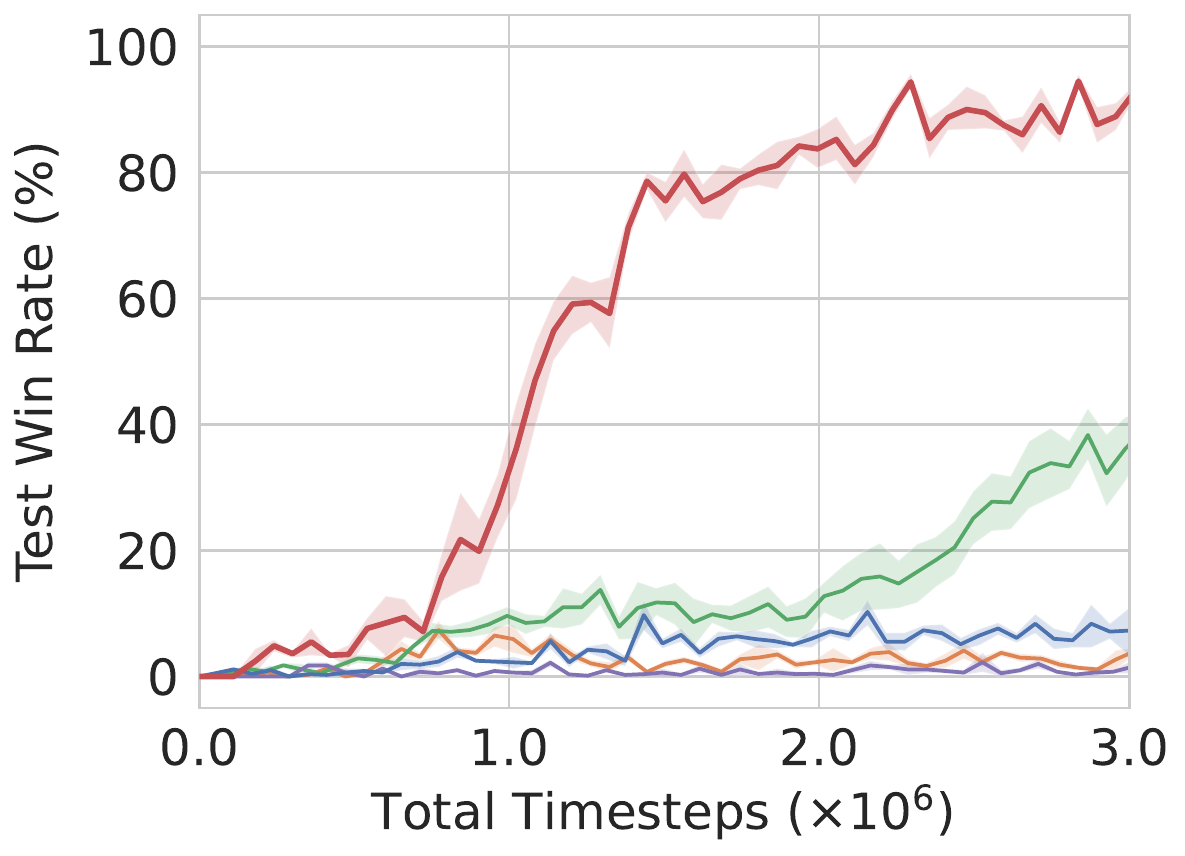}}\hfill
  \subfloat[Unseen Scale]{\includegraphics[width=0.333\textwidth]{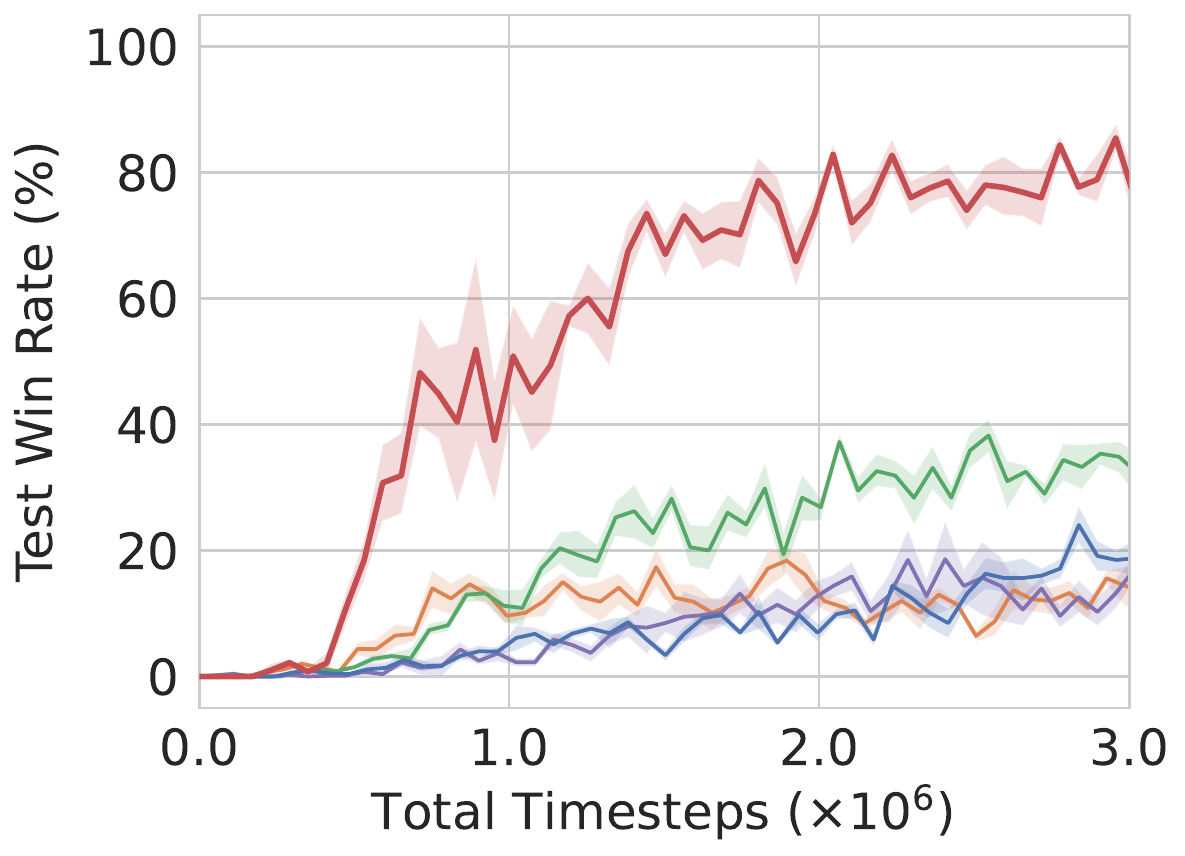}}\hfill
  \subfloat[Unseen C \& S]{\includegraphics[width=0.333\textwidth]{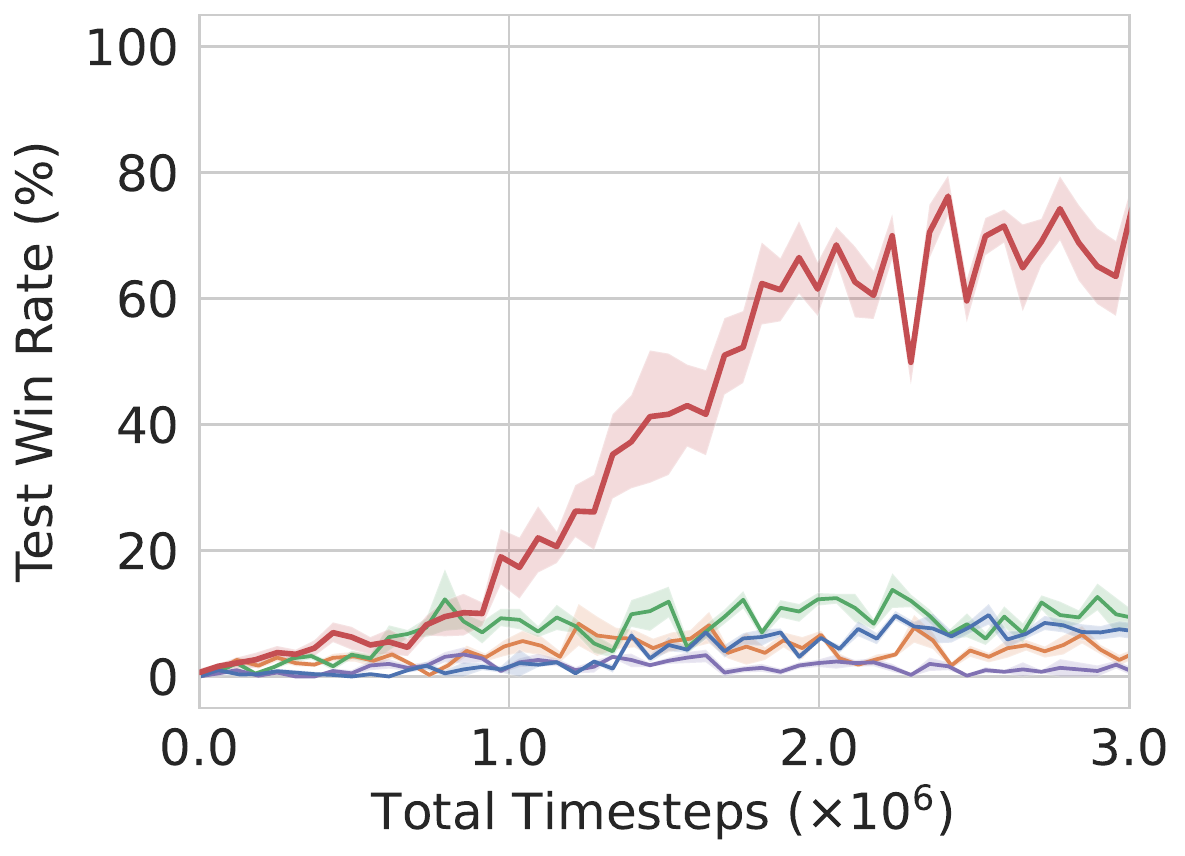}}
  \caption{Learning curves of our method and baselines in 3 multi-task Predator-Prey scenarios, where the results are provided by testing the trained agents on the unseen tasks.}
  \label{fig:multi-task-prey}
  \vspace{-0.2cm}
\end{figure*}

\begin{table}[!t]
  \centering
  \caption{The area under curve and test win rate of our method and baselines in 3 multi-task Predator-Prey scenarios, where the results are provided by testing the trained agents on the unseen tasks.}
  \label{tab:multi-task-prey}
  \resizebox{0.48\textwidth}{!}{%
  \begin{tabular}{@{}cccc@{}}
  \toprule
  \multirow{1}{*}{\textbf{Method}} & \makecell{\textbf{Unseen} \\  \textbf{Capability} } & \makecell{\textbf{Unseen} \\ \textbf{Scale}}    & \makecell{\textbf{Unseen}  \\ \textbf{C \& S}}    \\ \midrule
  &\multicolumn{3}{c}{\textbf{Area Under Curve}} \\ \cmidrule(l){2-4} 
  \textbf{Attn (VDN)}~\cite{VDN}  & 0.03 $\pm$ 0.00  & 0.10 $\pm$ 0.02  & 0.04 $\pm$ 0.01 \\ \specialrule{0em}{1pt}{1pt}
  \textbf{Attn (QMIX)}~\cite{QMIX}  & 0.13 $\pm$ 0.06  & 0.20 $\pm$ 0.03  & 0.08 $\pm$ 0.02 \\ \specialrule{0em}{1pt}{1pt}
  \textbf{REFIL (VDN)}~\cite{REFIL}  & 0.01 $\pm$ 0.00  & 0.08 $\pm$ 0.04  & 0.01 $\pm$ 0.01 \\ \specialrule{0em}{1pt}{1pt}
  \textbf{REFIL (QMIX)}~\cite{REFIL}  & 0.04 $\pm$ 0.01  & 0.08 $\pm$ 0.01  & 0.04 $\pm$ 0.00 \\ \specialrule{0em}{1pt}{1pt}\midrule
  \textbf{Ours}  & \textbf{0.55 $\pm$ 0.07}  & \textbf{0.55 $\pm$ 0.08}  & \textbf{0.37 $\pm$ 0.09} \\ \specialrule{0em}{1pt}{1pt}\midrule 
  &\multicolumn{3}{c}{\textbf{Test Win Rate}} \\ \cmidrule(l){2-4} 
  \textbf{Attn (VDN)}~\cite{VDN}  & 0.05 $\pm$ 0.04  & 0.14 $\pm$ 0.03  & 0.05 $\pm$ 0.04 \\ \specialrule{0em}{1pt}{1pt}
  \textbf{Attn (QMIX)}~\cite{QMIX}  & 0.37 $\pm$ 0.13  & 0.33 $\pm$ 0.17  & 0.07 $\pm$ 0.02 \\ \specialrule{0em}{1pt}{1pt}
  \textbf{REFIL (VDN)}~\cite{REFIL}  & 0.01 $\pm$ 0.02  & 0.12 $\pm$ 0.07  & 0.01 $\pm$ 0.01 \\ \specialrule{0em}{1pt}{1pt}
  \textbf{REFIL (QMIX)}~\cite{REFIL}  & 0.08 $\pm$ 0.08  & 0.20 $\pm$ 0.12  & 0.08 $\pm$ 0.04 \\ \specialrule{0em}{1pt}{1pt}\midrule
  \textbf{Ours}  & \textbf{0.89 $\pm$ 0.05}  & \textbf{0.78 $\pm$ 0.07}  & \textbf{0.75 $\pm$ 0.06} \\ \specialrule{0em}{1pt}{1pt}\bottomrule 
  \end{tabular}}
  \vspace{-0.3cm}
\end{table}

\subsection{Multi-task Environments} 

To further evaluate the proposed method in the multi-task setting, we conduct experiments on 6 multi-task SMAC scenarios, including 3 basic scenarios (\emph{3-8csz}, \emph{3-8MMM}, \emph{3-8sz}) proposed by Iqbal and Sha~\cite{REFIL} and 3 new harder scenarios (\emph{5-11csz}, \emph{5-11MMM}, \emph{5-11sz}). The \emph{3-8sz} scenario contains 39 unique tasks of different scales, and the \emph{3-8MMM} and \emph{3-8csz} scenarios contain 66 tasks respectively. We follow the multi-task setting of Iqbal and Sha~\cite{REFIL} and design 3 new harder scenarios: (1) \emph{5-11csz} contains two symmetrical teams, each of which is composed of 5 to 8 Stalkers/Zealots and 0 to 3 Colossi, resulting in 120 unique tasks. (2) \emph{5-11sz} contains two symmetrical teams, each of which is composed of 5 to 11 Stalkers/Zealots, resulting in 63 unique tasks. (3) \emph{5-11MMM} contains two symmetrical teams, each of which is composed of 5 to 8 Marines/Marauders and 0 to 3 Medics, resulting in 120 unique tasks. According to the difficulty of training, we can also divide these scenarios into three categories: \textbf{Easy}~(\emph{3-8csz}, \emph{5-11csz}), \textbf{Hard}~(\emph{3-8MMM}, \emph{3-8sz}) and \textbf{Super Hard}~(\emph{5-11MMM}, \emph{5-11sz}).
We compare the proposed OPT with various multi-task baselines: VDN~\cite{VDN}, QMIX~\cite{QMIX}, REFIL~\cite{REFIL}. We realize the attention-based VDN and QMIX for the multi-task setting as in REFIL~\cite{REFIL}. The models train simultaneously on multiple tasks sampled uniformly at each episode. For more exploration, we extend the epsilon anneal time to 500K for all the compared methods.

The proposed method is particularly beneficial in this multi-task setting, which enables us to explore the diverse common interaction prototypes and generalize across various-scale tasks. Figure~\ref{fig:multi-task} and Table~\ref{tab:multi-task} report the experimental results of all comparison methods on 6 multi-task SMAC scenarios. The results show that our proposed method consistently outperforms the baselines, which demonstrates the effectiveness of interaction pattern disentangling. In the easy scenarios~(\emph{3-8csz}, \emph{5-11csz}), REFIL can also learn effective cooperation strategies based on its random grouping mechanism. However, the latent interaction patterns still remain entangled in REFIL, which may severely downgrade the convergence rate. Conversely, our proposed OPT allows for the explicit shareable pattern disentangling across tasks, improving the learning efficiency and performance of the agents. In the hard \emph{3-8MMM} scenario, OPT achieves the performances on par with REFIL and still maintains a higher learning efficiency. Moreover, in the more difficult \emph{3-8sz} and \emph{5-11MMM} scenarios, OPT demonstrates its advantages in the transferability of the common prototypes and achieves gratifying results. In the super hard \emph{5-11sz} scenario, the performance gap between OPT and REFIL is not obvious, but most baselines including Attn~(VDN), Attn~(QMIX), REFIL~(VDN) fail to learn any effective policy. To this end, OPT also successfully discovers several effective common interaction prototypes.

\begin{figure*}[!t]
  \centering
  \subfloat[Ablation study on three major components.]{\includegraphics[scale=0.33]{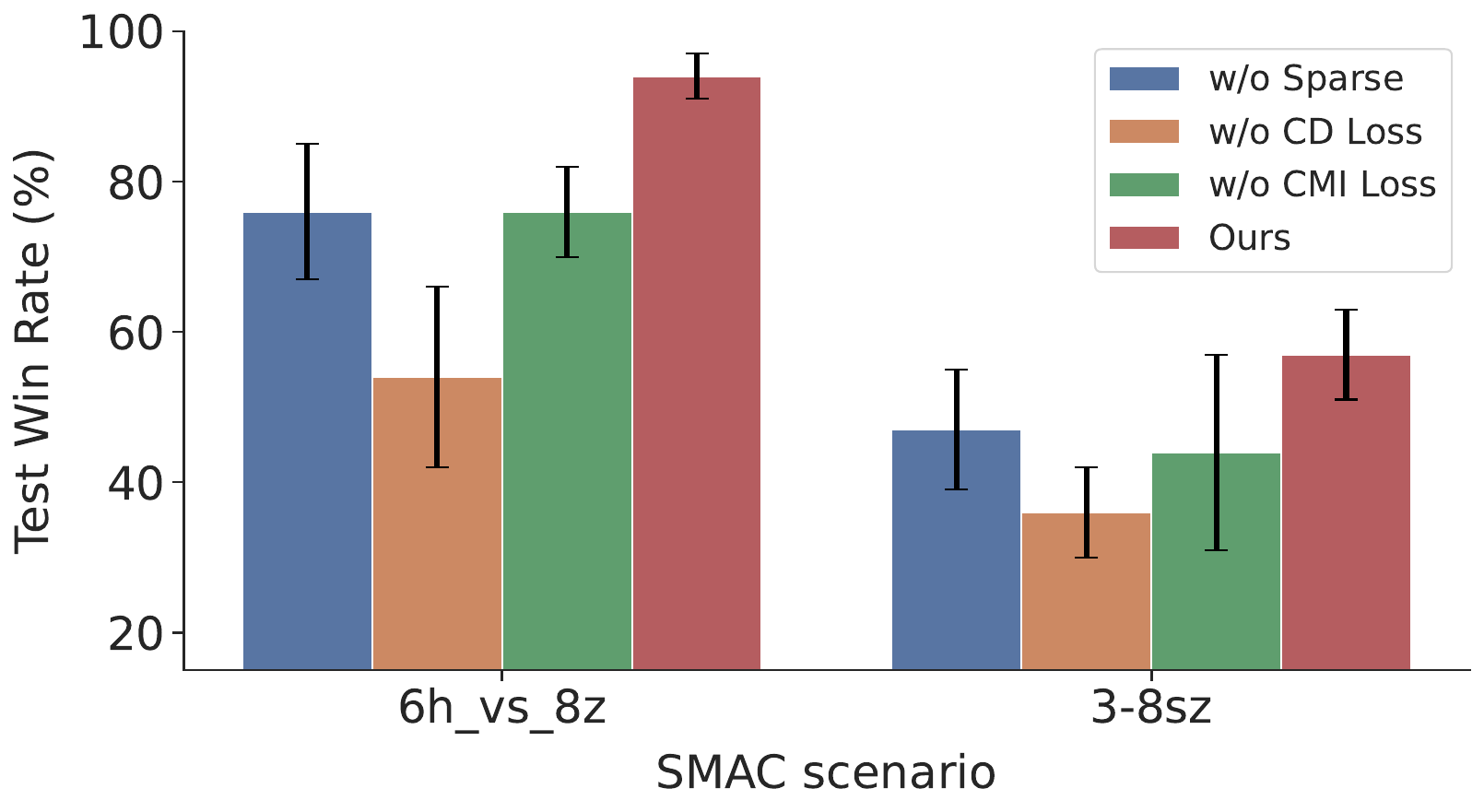}\label{fig:ablation1}}\hfill
  \subfloat[Ablation study on the interaction prototype number.]{\includegraphics[scale=0.33]{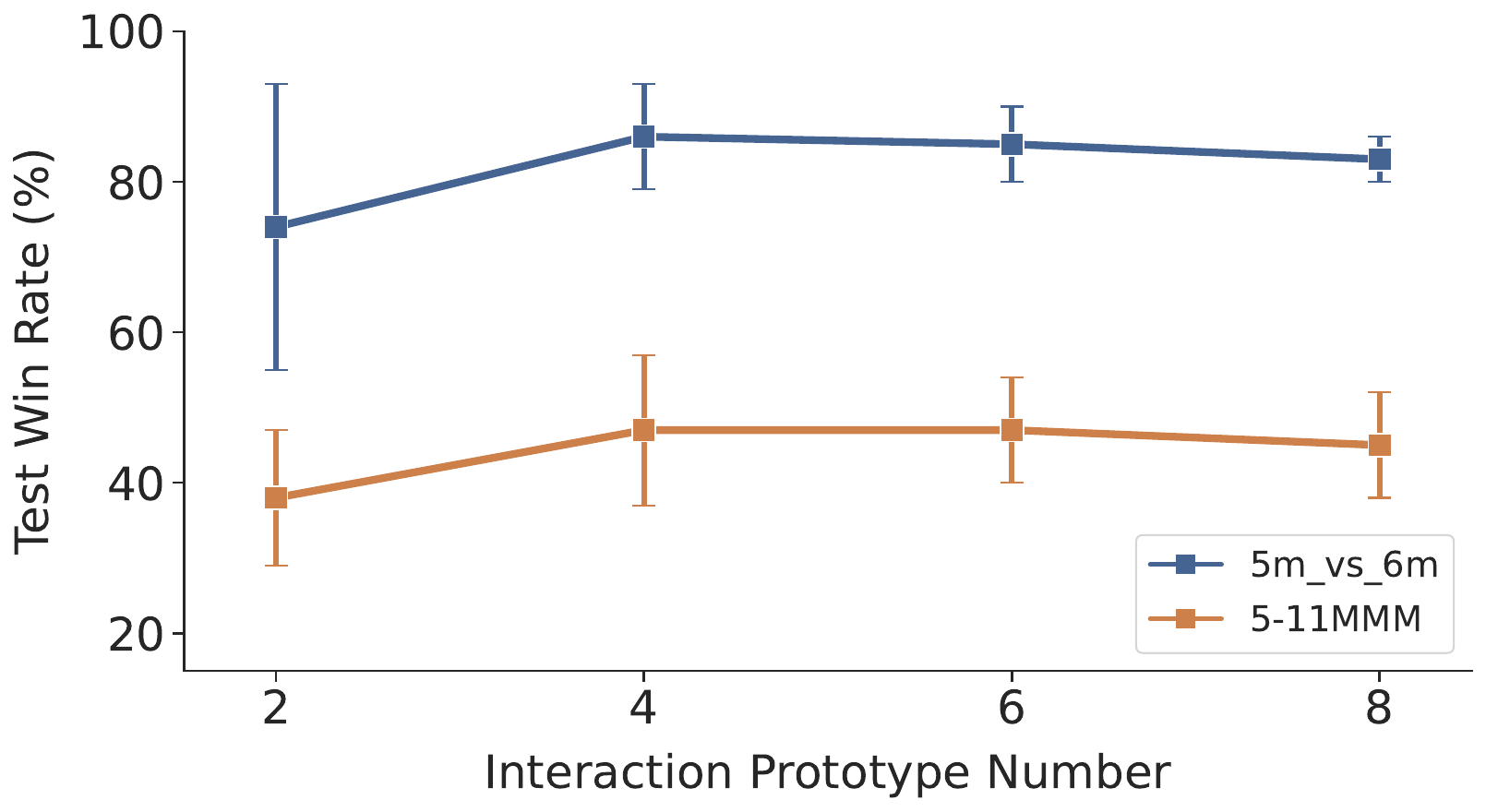}\label{fig:ablation2}}
  \caption{Comparing results over the ablations of OPT.}
  \label{fig:ablation}
\end{figure*}

\subsection{Zero-Shot Generalization Analysis}

We further design a multi-task Predator-Prey Game~\cite{mahajan2022generalization} to test the zero-shot generalization ability of OPT. Specifically, $N_{a}$ agents~(predators), $N_{p}$ prey and $N_{o}$ fixed obstacles are initialized in a grid world, where $\{N_{a}, N_{p}, N_{o}\}$ describes the scale of each task. Besides, each agent is equipped with an attack capability $C_{a}$, and each prey is equipped with a defense capability $C_{p}$. At each time step, agents with limited sight range can move, stop or capture, while prey randomly move and obstacles block movement. The goal of agents is to capture all prey, where one prey can be captured only when the total attack capability of agents is greater than or equal to the defense capability of the prey. The initialized scale and the capability are randomly sampled for each task during training. 
We conduct experiments on 3 multi-task Predator-Prey scenarios, where the agents are tested in the tasks with: (1) \emph{Unseen Capability}. (2) \emph{Unseen Scale}. (3) \emph{Unseen Capability \& Scale~(C \& S)}.

Figure~\ref{fig:multi-task-prey} and Table~\ref{tab:multi-task-prey} report the zero-shot testing results of all comparison methods on 3 multi-task Predator-Prey scenarios, where our OPT consistently outperforms baselines by a large margin. All multi-task baselines, even the state-of-the-art REFIL, perform poorly in the unseen tasks. The results empirically suggest that the random grouping strategy of REFIL is limited to adopting policies in the trained tasks. In contrast, OPT disentangles the intertwined entity interactions in an explicit way, which encourages agents to extract the shareable interaction prototypes that can be generalized to the unseen tasks.

\subsection{Ablations and Visualization}

To understand the superior performance of OPT, we carry out ablation studies to test the contribution of its three main components: sparsemax, contrastive disagreement~(CD) loss and conditional mutual information~(CMI) loss. Following methods are included in the evaluation: (i) OPT without sparsemax (denoted by \emph{w/o Sparse}); (ii) OPT without CD loss (denoted by \emph{w/o CD Loss}); (iii) OPT without CMI loss (denoted by \emph{w/o CMI Loss}). The results on both the single-task and multi-task scenarios are shown in Figure~\ref{fig:ablation1}. By comparing OPT without sparsemax and without CD loss, we can conclude that sparsity can help facilitate learning but the significant performance of OPT is mostly due to the restriction of CD loss. CD loss constrains the interaction prototypes to uniformly distribute on a hypersphere without divergence, which makes a major contribution to interaction pattern disentangling. Sparsemax mainly serves as an auxiliary means that constructs the more compact interaction prototypes. Moreover, CMI loss aims to stabilize the training process. When without CMI loss, the optimization results often derive a high standard deviation, especially in the multi-task scenario. This instability also leads to a significant drop in the performance.

\begin{figure*}[!t]
  \centering
  \begin{minipage}[t]{.335\linewidth}
    \subfloat[2c3s5z in 5-11csz]
      {\includegraphics[width=\linewidth]{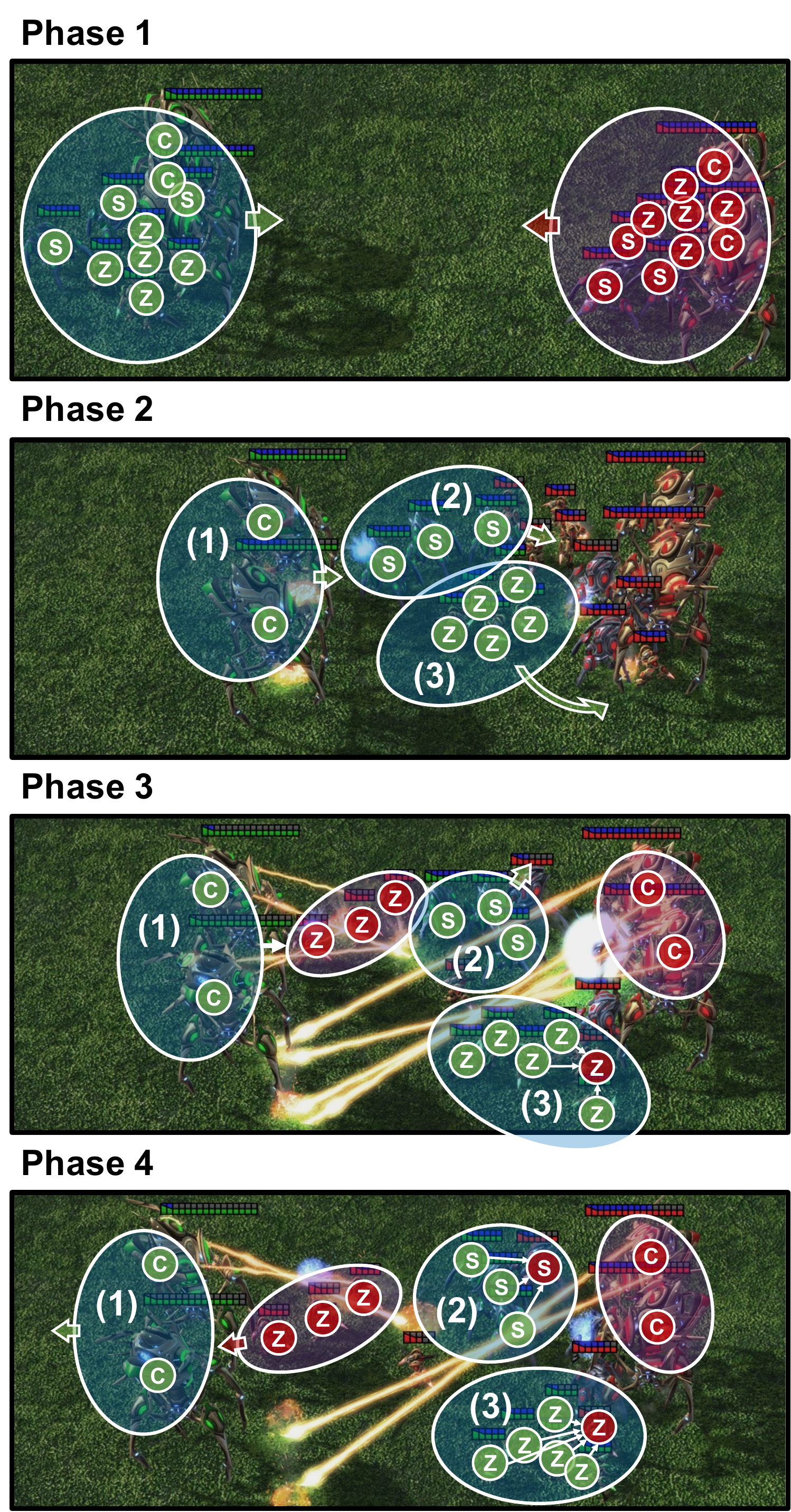}\label{fig:ex1}}%
  \end{minipage}\;
  \begin{minipage}[t]{.335\linewidth}
    \subfloat[1c4s4z in 5-11csz]
      {\includegraphics[width=\linewidth]{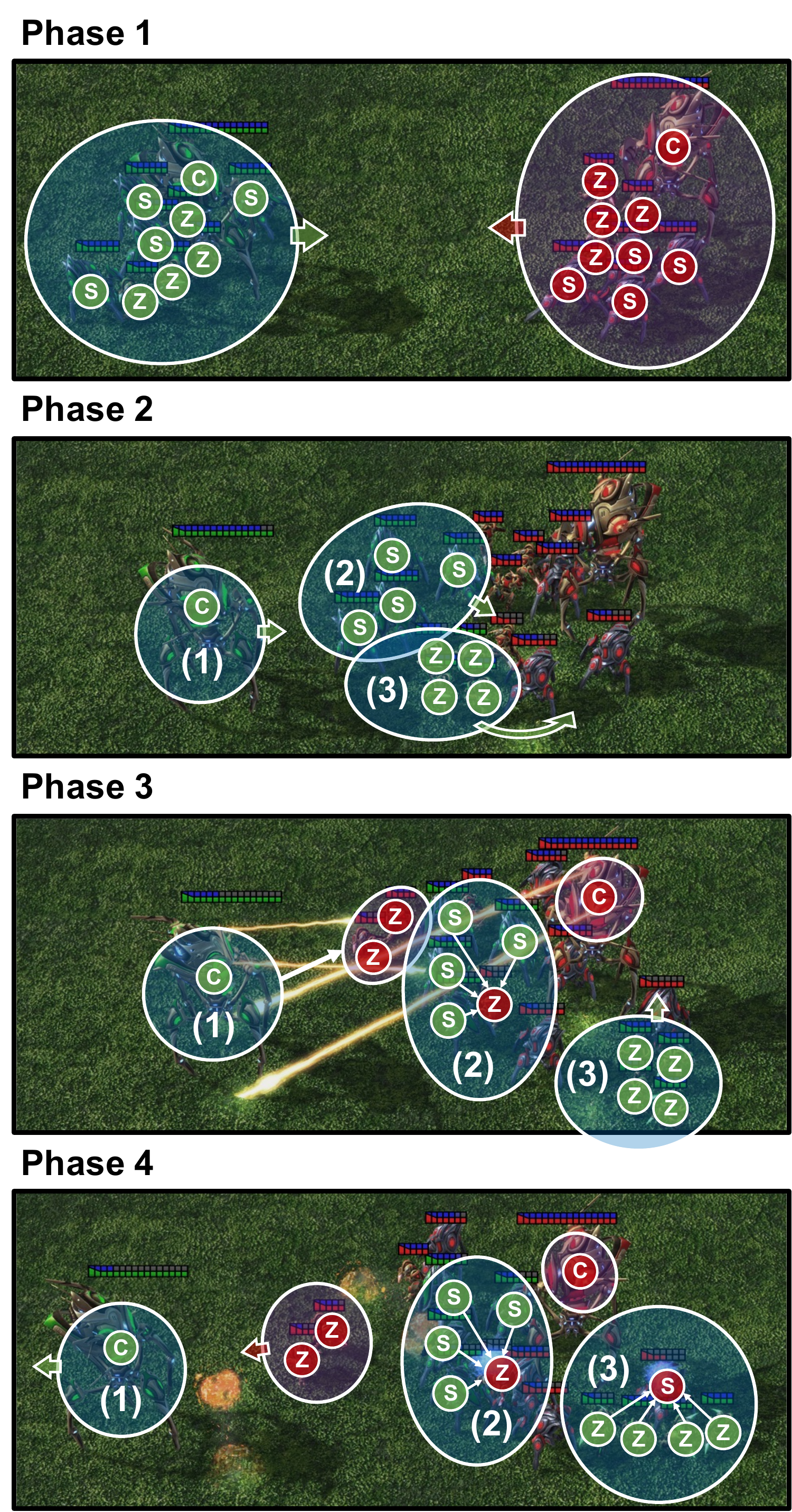}\label{fig:ex2}}%
  \end{minipage}\;
  \begin{minipage}[t]{.283\linewidth}
    \subfloat[Interaction prototypes of 2c3s5z]
      {\includegraphics[width=\linewidth]{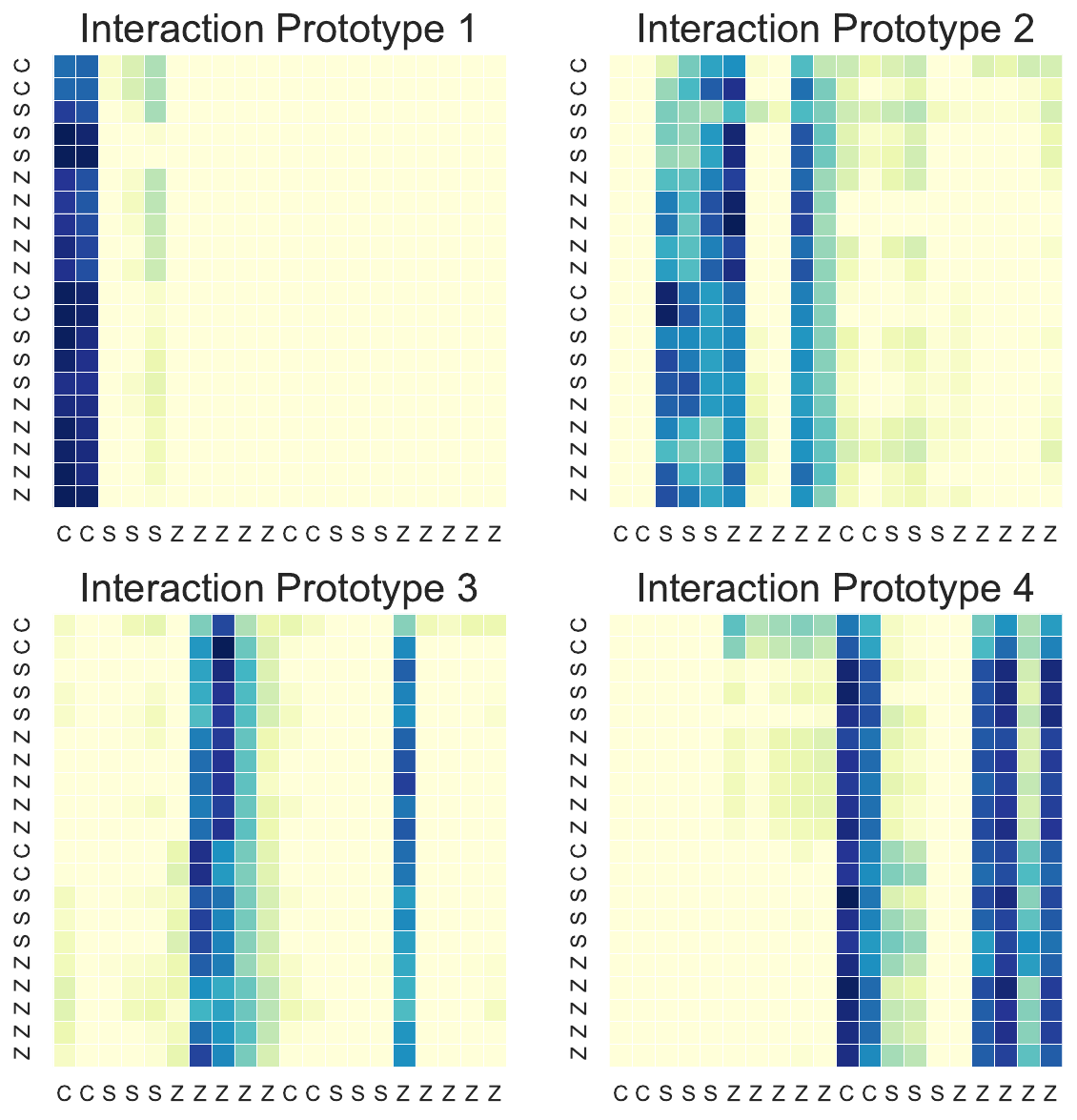}\label{fig:ex3}}\\
    \subfloat[Interaction prototypes of 1c4s4z]
      {\includegraphics[width=\linewidth]{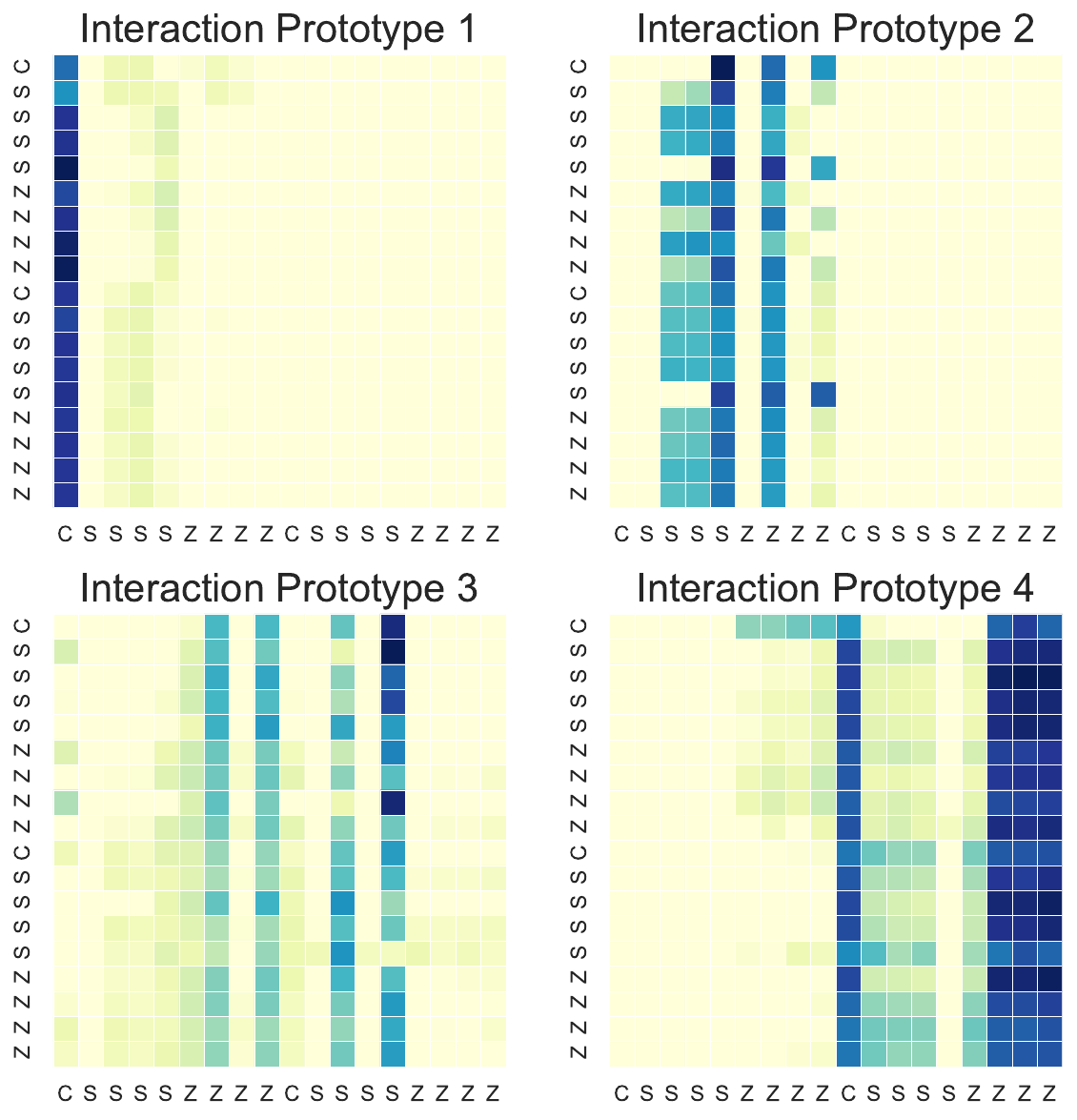}\label{fig:ex4}}
  \end{minipage}%
  \caption{A visualization example of the common interaction prototypes that emerge across tasks in the 5-11csz scenario. Please zoom in for a better view. (a) Each combat team of 2c3s5z is composed of 2 Colossi, 3 Stalkers and 5 Zealots. (b) Each combat team of 1c4s4z is composed of 1 Colossus, 4 Stalkers and 4 Zealots. Green and red shadows indicate the agent and enemy formations, respectively. Green and red arrows indicate the moving direction of the agent and enemy formations, respectively. White arrows indicate the attack action of the agents. (c) The interaction prototypes of 2c3s5z in the mixing network at phase 4. (d) The interaction prototypes of 1c4s4z in the mixing network at phase 4. The entities on the axes are arrayed by agents first and enemies later. 
  Diverse common tactics based on the common prototypes emerge in combat tasks with different scales: (1) Hit and Run based on Prototype 1\&4, \textit{i.e.}, drawing the fire, moving the agents away and fighting back again. (2) Coordinated Move and Force Fire based on Prototype 2, \textit{i.e.}, moving together and focusing fire on an enemy. (3) Sneak Attack based on Prototype 3, \textit{i.e.}, sneaking around behind the enemy and attacking the alone enemy.}
  \label{fig:ex}
\end{figure*}

Moreover, to study the impact of the interaction prototype number on the performance of OPT, we conduct an ablation study as shown in Figure~\ref{fig:ablation2}.
When the number of interaction prototypes increases from 2 to~4, the performance gain of OPT becomes larger. This result verifies the effectiveness of the interaction pattern disentangling. Different interaction prototypes still remain entangled when the number of interaction prototypes is small, resulting in a worse performance. However, as the number of interaction prototypes becomes too large, disentanglement will be more challenging, yielding lower performance gains. After reaching the peak at 4, the performance slightly drops, but in general, our method is not very sensitive when the number of interaction prototypes is large. More interestingly, we find that the standard deviation of the results becomes smaller as the number of interaction prototypes increases, which indicates that providing more interaction prototypes for restructuring enables the agents to stabilize the final policy.

To further explain the learned interaction prototypes from OPT, we conduct a qualitative analysis. As shown in Figure~\ref{fig:ex1} and~\ref{fig:ex2}, we find that OPT can effectively grasp diverse tactics and utilize these common tactics across various-scale tasks. At the beginning of the battle, two combat teams first get closer to each other. Then our combat team assembles the agents into formations based on their unit types. Since Colossus is a heavy support unit with high defense and a large shooting range, Colossi learn to draw the fire by kiting the enemies. Colossi make the enemies give chase while maintaining enough distance, so that little or no damage is incurred. Moreover, Stalkers often tend to move together and focus fire on an enemy due to their high attack damage. Most interestingly, Zealots master the ability of sneak attack, which bypasses the disadvantages of low defense and low attack damage. The visualization of the interaction prototypes in Figure~\ref{fig:ex3} and~\ref{fig:ex4}  shows that our proposed method successfully disentangles the interaction pattern into several interaction prototypes that are sparse and distinguishable. Moreover, the interaction prototypes between different tasks are highly similar, which verifies the generalization of these prototypes. The shareable interaction prototypes enable our agents to construct and reuse the effective policy, and finally improve the performance~of~the~model.

\section{Conclusion\label{sec:conclusion}}

In this work, we propose a novel interaction pattern disentangling approach, termed as OPT, that enables us to take advantage of the factorized interaction prototypes for multi-agent reinforcement learning. OPT follows a restructuring-by-disentangling scheme, where several distinguishable sparse prototypes are generated and then selectively reassembled together for the final focused policy. We validate OPT over the StarCraft II micromanagement, Google Research Football and Predator-Prey benchmarks, and showcase that it yields results significantly superior to the state-of-the-art techniques in both single-task and multi-task settings. 
To our best knowledge, this paper is the first attempt towards interaction pattern disentangling in MARL. 
In our future work, we will extend our method to design a more lightweight disentanglement framework for the large-scale problem, which is more challenging in terms of the exponential complexity as the number of agents increases.

\bibliographystyle{IEEEtran}
\bibliography{ref}

\newpage

\appendix[{Proof of the CMI Loss}]

We propose a conditional mutual information (CMI) objective in the utility network to stabilize the aggregation weights:
\begin{align*}
    I(\boldsymbol{\omega}^a_t ; \tau^a_{t-1} | o^a_t).
\end{align*}
Based on the variational inference, we introduce a variational approximator $q_{\psi}$ parameterised by $\psi$ to approximate the posterior $p(\boldsymbol{\omega}^a_t | \tau^a_{t-1}, o^a_t)$ and derive a tractable lower bound of the CMI objective:
\allowdisplaybreaks
\begin{align*}
    &I(\boldsymbol{\omega}^a_t ; \tau^a_{t-1} | o^a_t) 
    \\=\;&H(\boldsymbol{\omega}^a_t | o^a_t) - H(\boldsymbol{\omega}^a_t | \tau^a_{t-1}, o^a_t)
    \\=\;&H(\boldsymbol{\omega}^a_t | o^a_t) + \mathbb{E}_{\tau^a_{t-1}, o^a_t} \Bigl[\mathbb{E}_{{\omega}^a_t} \bigl[\log p(\boldsymbol{\omega}^a_t | \tau^a_{t-1}, o^a_t) \bigl] \Bigl]
    \\=\;&H(\boldsymbol{\omega}^a_t | o^a_t) + \mathbb{E}_{\tau^a_{t-1}, o^a_t} \Bigl[\mathbb{E}_{{\omega}^a_t} \bigl[\log p(\boldsymbol{\omega}^a_t | \tau^a_{t-1}, o^a_t) \\ &  - \log q_{\psi }(\boldsymbol{\omega}^a_t | \tau^a_{t-1}, o^a_t) + \log q_{\psi }(\boldsymbol{\omega}^a_t | \tau^a_{t-1}, o^a_t)  \bigl] \Bigl]
    \\=\;&H(\boldsymbol{\omega}^a_t | o^a_t) + \mathbb{E}_{\tau^a_{t-1}, o^a_t} \Bigl[\mathbb{E}_{{\omega}^a_t} \bigl[\log q_{\psi }(\boldsymbol{\omega}^a_t | \tau^a_{t-1}, o^a_t)  \bigl] \Bigl] \\ &+ \mathbb{E}_{\tau^a_{t-1}, o^a_t} \Bigl[ \operatorname{KL}\bigl[p(\boldsymbol{\omega}^a_t | \tau^a_{t-1}, o^a_t) \parallel q_{\psi }(\boldsymbol{\omega}^a_t | \tau^a_{t-1}, o^a_t)\bigl] \Bigl].
\end{align*} 
Due to the non-negativity of the KL divergence, we can obtain the lower bound of the CMI objective as
\begin{align*}
    &I(\boldsymbol{\omega}^a_t ; \tau^a_{t-1} | o^a_t) 
    \\ \geq\;&H(\boldsymbol{\omega}^a_t | o^a_t) + \mathbb{E}_{\tau^a_{t-1}, o^a_t} \Bigl[\mathbb{E}_{{\omega}^a_t} \bigl[\log q_{\psi }(\boldsymbol{\omega}^a_t | \tau^a_{t-1}, o^a_t)  \bigl] \Bigl].
\end{align*}
Then we further simplify this lower bound as follows:
\begin{align*}
    &H(\boldsymbol{\omega}^a_t | o^a_t) + \mathbb{E}_{\tau^a_{t-1}, o^a_t} \Bigl[\mathbb{E}_{{\omega}^a_t} \bigl[\log q_{\psi }(\boldsymbol{\omega}^a_t | \tau^a_{t-1}, o^a_t)  \bigl] \Bigl]
    \\=\;&H(\boldsymbol{\omega}^a_t | o^a_t) + \mathbb{E}_{\tau^a_{t-1}, o^a_t}  \Bigl[ \int p(\boldsymbol{\omega}^a_t | o^a_t) \log q_{\psi }(\boldsymbol{\omega}^a_t | \tau^a_{t-1}, o^a_t)  \Bigl]
    \\=\;&H(\boldsymbol{\omega}^a_t | o^a_t) - \mathbb{E}_{\tau^a_{t-1}, o^a_t} \Bigl[ H \bigl( p(\boldsymbol{\omega}^a_t | o^a_t), q_{\psi }(\boldsymbol{\omega}^a_t | \tau^a_{t-1}, o^a_t) \bigl) \Bigl]
    \\=\;&-\mathbb{E}_{\tau^a_{t-1},o^a_t}  \Bigl[ \operatorname{KL}\bigl[p(\boldsymbol{\omega}^a_t | o^a_t) \parallel q_{\psi }(\boldsymbol{\omega}^a_t | \tau^a_{t-1}, o^a_t)\bigl]   \Bigl].
\end{align*}
Therefore, our final CMI objective can be formalized as a loss function:
\begin{align*}
  \mathcal{L}_{CMI}(\theta,\phi,\psi) = \mathbb{E}_{\mathcal{D}} \Bigl[ \operatorname{KL}\bigl[p_{\phi }(\boldsymbol{\omega}^a_t | \bar{x}_t^a) \parallel q_{\psi }(\boldsymbol{\omega}^a_t | h^a_{t-1}, \bar{x}_t^a)\bigl]\Bigl],
\end{align*}
where $\operatorname{KL}(\cdot)$ is the KL divergence function, $\bar{x}_t^a$ is the input embedding of the observation $o^a_t$ after mean pooling operation, and a GRU network is used to encode the history trajectory $\tau^a_{t-1}$ into $h^a_{t-1}$.

\end{document}